\documentclass{article}

\usepackage[utf8]{inputenc} % allow utf-8 input
\usepackage[T1]{fontenc}    % use 8-bit T1 fonts
\usepackage{hyperref}       % hyperlinks
\usepackage{url}            % simple URL typesetting
\usepackage{booktabs}       % professional-quality tables
\usepackage{amsfonts}       % blackboard math symbols
\usepackage{nicefrac}       % compact symbols for 1/2, etc.
\usepackage{microtype}      % microtypography

\usepackage{amssymb,amsfonts,amsmath,amscd,dsfont,mathrsfs}
\usepackage{pgf,pgfarrows,pgfnodes}
\usepackage{multicol, algorithmic,multirow}

\usepackage{caption}
\usepackage{subcaption}
\usepackage{tikz}
\usepackage{lipsum}

\usepackage{color}
\usepackage[numbers]{natbib}
\usepackage{amsmath,amsfonts, amssymb}
\usepackage{amsthm}
\usepackage{mathrsfs}

\usepackage{dsfont} % for mathds
\usepackage{xspace}

\newtheorem{lemma}{Lemma}
\newtheorem{theorem}{Theorem}
\newtheorem{corollary}{Corollary}
\newtheorem{proposition}{Proposition}
\newtheorem{remark}{Remark}
\newtheorem{assume}{Assumption}

\newcommand{\E}{\mathbb{E}}

\newcommand{\prob}[1]{ \mathbb{P}\left(#1 \right) }

\newcommand{\define}{\triangleq}

\newcommand{\reals}{\mathbb{R}}
\newcommand{\naturals}{\mathbb{N}}
\newcommand{\ones}{\mathds{1}}
\newcommand{\identy}{\mathbf{I}}

\newcommand{\abs}[1]{\left| #1 \right|}
\newcommand{\norm}[1]{\left\lVert#1\right\rVert}
\newcommand{\lnorm}[2]{\norm{#1}_{#2}}
\newcommand{\maxnorm}[1]{\vertiii{#1}_{\infty}}
\newcommand{\onorm}[1]{\vertiii{#1}_{1}}
\newcommand{\tnorm}[1]{\vertiii{#1}_{2}}

\newcommand{\vopsymb}{\circ}
\newcommand{\vop}[2]{#1 \vopsymb #2}
\newcommand{\vect}[1]{\overline{#1}}

\newcommand{\la}{\langle}
\newcommand{\ra}{\rangle}
\newcommand{\cF}{\mathcal{F}}
\newcommand{\cV}{\mathcal{V}}

\newcommand{\cG}{\mathcal{G}}
\newcommand{\vcF}{\overline{\cF}}

\newcommand{\cX}{\mathcal{X}}
\newcommand{\cU}{\mathcal{U}}

\newcommand{\ty}{\widetilde{y}}
\newcommand{\tY}{\widetilde{Y}}
\newcommand{\D}{D}
\newcommand{\tD}{\widetilde{D}}

\newcommand{\ny}{m}

\newcommand{\nndfp}[4]{d_{#1, #2} (#3, #4 )}

\newcommand{\nnd}[2]{\nndfp{\cF}{\phi}{#1}{#2}}
\newcommand{\nndff}[3]{d_{#1} (#2, #3 )}
\newcommand{\expectD}[2]{\underset{#1}{\mathbb{E}}\left[ #2 \right]}
\newcommand{\smallexpectD}[2]{\underset{#1}{\mathbb{E}}\big[ #2 \big]}

\newcommand{\x}[1]{#1_{X}}
\newcommand{\ygx}[1]{#1_{Y|X=x}}
\newcommand{\tygx}[1]{#1_{\tY|X=x}}
\newcommand{\xy}[1]{#1_{X, Y}}
\newcommand{\xty}[1]{#1_{X, \tY}}

\newcommand{\vygx}[1]{\ygx{\vect{#1}}}
\newcommand{\vygxP}{\vygx{P}}

\newcommand{\vtygx}[1]{\tygx{\vect{#1}}}
\newcommand{\vtygxP}{\vtygx{\tP}}

\newcommand{\tP}{\widetilde{P}}
\newcommand{\tQ}{\widetilde{Q}}
\newcommand{\htP}{\widetilde{P}_n}
\newcommand{\htQ}{\widetilde{Q}_n}

\newcommand{\vp}{\vect{p}}
\newcommand{\vq}{\vect{q}}

\newcommand{\vertiii}[1]{{\vert\kern-0.25ex\vert\kern-0.25ex\vert #1 
    \vert\kern-0.25ex\vert\kern-0.25ex\vert}}

\newcommand{\tv}[2]{d_{\rm TV}\left(#1, #2\right)}
\usepackage{mathtools}
\DeclarePairedDelimiterX{\infdivx}[2]{(}{)}{%
	#1\;\delimsize\|\;#2%
}
\newcommand{\kl}{d_{\rm KL}\infdivx*}
\newcommand{\js}{d_{\rm JS}\infdivx*}

\newcommand{\confus}{C}
\newcommand{\normconfus}{\vertiii{\confus^{-1}}_\infty^{-1}}
\newcommand{\normconfust}{\maxnorm{\confus^{-1}}^{-2}}
\newcommand{\normconfusinv}{\vertiii{\confus^{-1}}_\infty}
\newcommand{\NND}{neural network distance\xspace}

\renewcommand\footnotemark{}

\title{Robustness of Conditional GANs to Noisy Labels}

\author{
Kiran Koshy Thekumparampil$^\dagger$,   Ashish Khetan$^\dagger$, Zinan Lin$^\ddagger$, Sewoong Oh$^\dagger$
\thanks{Author emails are {\text thekump2@illinois.edu}, \text{khetan2@illinois.edu}, {\text{zinanl@andrew.cmu.edu}, and \text{swoh@illinois.edu}}. 
This work used the Extreme Science and Engineering Discovery Environment (XSEDE), which is supported by National Science Foundation grant number OCI-1053575.  Specifically, it used the Bridges system, which is supported by NSF award number ACI-1445606, at the Pittsburgh Supercomputing Center (PSC).}\\
$^\dagger$University of Illinois at Urbana-Champaign,
$^\dagger$Carnegie Mellon University \\
}

\date{}

\begin{document}

\maketitle

\begin{abstract}
We study the problem of learning conditional generators from noisy labeled samples, where 
the labels are corrupted by random noise.
A standard training of conditional GANs will not only produce samples with wrong labels, but also  generate poor quality samples. 
We consider two scenarios, depending on whether the noise model is known or not.
When the distribution of the noise is  known, 
we introduce a novel architecture which we call Robust Conditional GAN (RCGAN). 
The main idea is to corrupt the label of the generated sample before feeding to the adversarial discriminator, 
forcing the generator to 
produce samples with clean labels. 
This approach of passing through a matching noisy channel is 
justified by 
corresponding  multiplicative approximation bounds between the loss of the RCGAN and the distance between the clean real distribution and the generator distribution. 
This shows that the proposed approach is robust, 
when used with a carefully chosen discriminator architecture, known as projection discriminator. 
%performance depends on the noise statistics.  
%and provide sample complexity of the loss in neural network distances under standard assumptions on the discriminator class. 
%Our analyses provides guidelines for the architecture design of RCGAN. 
%We give theoretical justification of our architectural choices. 
When the distribution of the noise is not known, we provide an extension of 
our architecture, which we call RCGAN-U, 
that learns the noise model simultaneously while training the generator. 
We show experimentally on MNIST and CIFAR-$10$ datasets that 
both the approaches consistently improve upon baseline approaches, 
and RCGAN-U  closely matches the performance of RCGAN. 
\end{abstract}

%%%%%%%%%%%%%%%%%%%%%%%%%%%%%%%%%%%%%%%%%%%%%%%%%%%%%%%%%%%%%%%%%%%%%%
% Introduction
%%%%%%%%%%%%%%%%%%%%%%%%%%%%%%%%%%%%%%%%%%%%%%%%%%%%%%%%%%%%%%%%%%%%%%

\section{Introduction}
\label{sec:intro} 

\label{sec:noise-model}

\label{sec:notations}

%\cite{miyato2018cgans}

% we study conditional GANs, which assume/rely on fidelity of the labels.
Conditional generative adversarial networks (GAN) have been widely successful in several applications including  
improving image quality, 
semi-supervised learning, 
reinforcement learning, 
category transformation, 
style transfer, 
image de-noising, 
compression,  
in-painting, and  
super-resolution \cite{mirza2014conditional,denton2015deep,VNK16,OOS16,ledig2016photo,zhu2017unpaired}. 
The goal of training a conditional GAN is to generate samples from 
distributions satisfying certain {\em conditioning} on some correlated features. 
Concretely,  given samples from joint distribution of a data point $x$ and a label $y$, 
we  want to learn to generate samples from the true conditional distribution of the real data  $P_{X | Y}$. 
A canonical conditional GAN studied in literature is the case of discrete label $y$ \cite{mirza2014conditional,OOS16,NYB16,miyato2018cgans}. 
Significant progresses have been made in this setting, which are typically evaluated on the quality of the conditional samples.  
These include measuring 
inception scores and intra Fr\'echet inception distances, 
visual inspection on downstream tasks such as category morphing and super resolution \cite{miyato2018cgans}, and  
faithfulness of the samples as measured by how accurately we can infer the class that generated the sample  \cite{OOS16}. 

We study the problem of training conditional GANs with noisy discrete labels. 
By noisy labels, we refer to a setting where the label $y$ for each example in the training set is randomly  corrupted.  
Such noise can result from an adversary deliberately corrupting the data \cite{biggio2011support} 
or from human errors in crowdsourced label collection \cite{dawid1979maximum,karger2011iterative}. 
% crowdsourced labels can be noisy
%significant literature on how to mitigate it, but require budget, assumptions, etc. depending on how much budget you have, can be unreliable.  
This can be modeled as a random process, where 
%\subsection{Noisy Label model}
%\label{sec:noise-model}
a clean data point $x\in \cX$ and its label $y\in[\ny]$ 
are drawn from a joint distribution $P_{X,Y}$ with $\ny$ classes. 
For each data point, the label is corrupted by passing through a noisy channel 
represented by a row-stochastic {\em confusion matrix} 
$C\in\reals^{\ny\times \ny}$ defined as 
$C_{ij} \triangleq {\mathbb P}(\tY = j | Y=i)$. 
This defines a joint distribution for the data point $x$ and a noisy label $\ty$: $\tP_{X,\tY}$. 
%The we define $\tP$ as the modified distribution of the random vector $(X, \tY) \in \cX \times [\ny]$, where $\tY$ is generated by flipping the original label $Y$ to $\tY$ according to the conditional probability distribution, 
%\begin{align}
%\prob{\tY = j | Y=i} = \confus_{ij} \; \forall i,j \in [\ny]\,.
%\end{align}
%We call the row(right)-stochastic matrix $\confus \in \reals^{\ny \times \ny}$, the {\it confusion matrix}. Notice that since rows sum to $1$, $\maxnorm\confus = \max_i \sum_j \abs{\confus_{ij}} = 1$.
%\subsection{Ambient GAN for noisy labels}
If we train a standard conditional GAN on noisy samples, then it solves the following optimization: 
\begin{align}
\min_{G \in \cG} \max_{\D \in \cF} V(G, D) = \expectD{(x,\ty) \sim \tP_{X,\tY}}{\phi\left(\D(x,\ty)\right)} + \expectD{z \sim N\,, y \sim \tP_{\tY}}{\phi\left(1- \D(G(z;y), y)\right)} \label{eq:bias_gan}
\end{align}
where $\phi$ is a  function of choice, 
$D$ and $G$ are the discriminator and the generator respectively 
optimized over  function classes $\cG$ and $\cF$ of our choice,  
and  $N$ is the distribution of the latent random vector. 
For typical choices of $\phi$, for example $\log(\cdot)$, 
and large enough function  classes $\cG$ and $\cF$, 
the optimal conditional generator learns to generate samples from 
$\tP_{X|\tY}$, the corrupted conditional distribution.  
In other words, it generates samples $X$ from classes other than what it is conditioned on.  
%If we are given noisy samples $(X, \tY)$ generated from the noise model in Section \ref{sec:noise-model}, the naive training of conditional GAN on these noisy samples, if successful, would only generate samples from the distribution of $X | \tY=\ty$, which is a biased version of the true conditional distribution $X | Y=y$. 
As the learned distribution exhibits  such a  bias, we call this naive approach the {\em Biased GAN}. 
% we study robustness of existing conditional GANs against,
Under this setting, there is a fundamental question of interest: 
can we design a novel conditional GAN that can generate 
samples from the true conditional distribution $P_{X|Y}$, 
even when trained on noisy samples? 
%
%\begin{enumerate}
%\item Noisy labeled model
%\item Ambient GAN (Noise Discriminator)
%\item Unbiased Discriminator is practically unstable.
%\end{enumerate}

Several aspects of this problem make it challenging and interesting. 
First, the performance of such robust GAN should depend on how noisy the channel $C$ is. 
If $C$ is rank-deficient, for instance, then there are multiple distributions that 
result in the same distribution after the corruption, and hence no 
reliable learning of the true distribution is possible.  We would ideally want 
a theoretical guarantee that shows such trade-off between $C$ 
and the robustness of GANs. 
Next, when the noise is from errors in crowdsourced labels, 
we might have some access to the confusion matrix $C$ from historical data. 
On  other cases of adversarial corruption, 
we might not have any information of $C$. 
We want to provide robust solutions to both. 
Finally, an important practical challenge in this setting is to correct the noisy labels in the training data. 
We address all such variations in our approaches and make the following contributions. 

\bigskip
\noindent
{\bf Our contributions.}  
We introduce two  architectures to 
train conditional GANs with noisy samples.  

First, when we have the knowledge of 
the confusion matrix $C$, we propose 
RCGAN (Robust Conditional GAN) in Section \ref{sec:rcgan}.  
We first prove that minimizing the RCGAN 
loss provably recovers the clean distribution $P_{X|Y}$
 (Theorem \ref{thm:gen_ub_lb}), 
under certain conditions on 
the class $\cF$ of discriminators 
we optimize over (Assumption \ref{eq:mix_inv}).   
We  show that such a condition on $\cF$ is also necessary, as without it, 
the training loss can be arbitrarily small while 
the generated distribution can be far from the real 
 (Theorem \ref{thm:gen_counter}). 
The assumption leads to our particular choice of the discriminator in 
RCGAN, called {\em projection discriminator} \cite{miyato2018cgans} that satisfies all the conditions (Remark \ref{rem:proj}). 
Finally, we  provide  
a finite sample generalization bound showing that  
the loss minimized in training RCGAN does generalize, and results in the learned distribution 
being close to the clean conditional distribution $P_{X|Y}$ (Theorem \ref{thm:nn_sample}). 
Experimental results in benchmark datasets confirm that 
RCGAN is robust against noisy samples, and improves significantly over 
the naive Biased GAN. 

Secondly, when we do not have access to $C$, 
we propose RCGAN-U (RCGAN with Unknown noise distribution) in Section \ref{sec:RCGAN-U}. 
We provide experimental results showing that performance gains similar to that of RCGAN can be achieved. 
%The main idea is to jointly learn the confusion matrix, while regularizing 
%those examples generated from the same class to be coherent. 
%This is ensures by adding two extra regularizers, 
%one for penalizing overlapping classes in the generated distribution, and 
%another penalizing mixing in the class labels. 
Finally, we showcase the practical use of thus learned conditional GANs, by
using it to fix the noisy labels in the training data. 
Numerical experiments confirm that 
the RCGAN framework provides a  more robust approach 
to correcting the noisy labels, compared to the state-of-the-art methods that rely only on 
discriminators.

\bigskip
\noindent
{\bf Related work.} 
Two popular training methods for generative models are 
variational auto-encoders \cite{kingma2013auto} 
and adversarial training \cite{goodfellow2014generative}. 
The adversarial training approach has made 
significant advances in several applications of practical interest.  
\cite{radford2015unsupervised,arjovsky2017wasserstein,berthelot2017began} 
propose new architectures that significantly improve the training in practical image datasets. 
\cite{zhu2017unpaired,isola2017image} 
propose new architectures to transfer the style of one image to the other domain. 
\cite{ledig2016photo,shrivastava2017learning} show 
how to enhance a given image with learned generator, by enhancing the resolution or 
making it more realistic.  
\cite{liang2017dual, 
vondrick2016generating} show how to generate videos and 
\cite{wu2016learning,
achlioptas2017representation} demonstrate that 3-dimensional models can be generated from adversarial training. 
\cite{causal} proposes a new architecture encoding causal structures in conditional GANs. 
\cite{mimic} introduces the state-of-the-art 
conditional independence tester.  
On a different direction, several recent approaches showcase how 
the manifold learned by the adversarial training can be used to solve inverse problems 
\cite{bora2017compressed,zhu2016generative,yeh2016semantic,van2018compressed}.

Conditional GANs have been proposed as a successful tool for various applications, including class conditional image generation \cite{OOS16}, image to image translation \cite{kim2017learning}, and image generation from text \cite{reed2016generative, zhang2017stackgan}. Most of the conditional GANs incorporate the class information by naively concatenating it to the input or feature vector at some middle layer \cite{mirza2014conditional, denton2015deep, reed2016generative, zhang2017stackgan}. AC-GANs \cite{OOS16} creates an auxiliary classifier to incorporate class information. Projection discriminator GAN \cite{miyato2018cgans}  takes an inner product between the embedded class vector and the feature vector. A recent work \cite{miyato2018spectral}  which proposes spectral normalization shows that high quality image generation on $1000$-class ILSVRC2012 dataset \cite {russakovsky2015imagenet} can be achieved using projection conditional discriminator.

Robustness of (unconditional) GANs against 
adversarial or random noise has 
recently been studied in \cite{bora2018ambientgan,xu2018robust}.
%{,schmidt2018adversarially,huang2018visual}. %gadelha20163d
\cite{xu2018robust} studies an adversary attacking the output of the discriminator, 
perturbing the discriminator output with random noise. 
The proposed architecture of RCGAN is inspired by a closely related work of AmbientGAN in \cite{bora2018ambientgan}. 
%The main idea is to corrupt the labels of the generated samples, let discriminator 
%classify based on those corrupted examples from both real and fake data. This is inspired by Ambient GAN \cite{bora2018ambientgan} 
%where a similar approach was proposed for unconditional GANs. 
AmbientGAN is a  general framework addressing any corruption on the data itself (not necessarily just the labels). 
Given a corrupted samples with known corruption, 
AmbientGAN applies that corruption to the output of the generator before feeding them to the discriminator. 
This has shown to successfully de-noise images in several practical scenarios. 

Motivated by the success of AmbientGAN in de-noising, we propose RCGAN. 
An important distinction is that 
we make specific architectural choices guided by 
our theoretical analysis that gives a significant gain in practice as shown in Section~\ref{sec:ambient}. 
% add novel regularizers that improves the performance significantly, 
Under the scenario of interest with noisy labels, we provide sharp analyses for both 
the population loss and the finite sample loss. 
Such sharp characterizations do not exist for  
the more general AmbientGAN scenarios. 
Further, our RCGAN-U does not require the knowledge of the confusion matrix,  departing from the AmbientGAN approach. 
% studies adversarial perturbation on the training data $X$, 
%with a known perturbation mechanism. 
%\cite{schmidt2018adversarially,huang2018visual}
%noisy labels/label correction - Learning with Noisy Labels \cite{natarajan2013learning}, 
Training classifiers from noisy labels is a closely related problem. 
Recently, \cite{natarajan2013learning, khetan2017learning} proposed a theoretically motivated classifier which minimizes the modified loss in presence of noisy labels and showed improvement over the robust classifiers \cite{liu2003building, stempfel2007learning, stempfel2009learning}.

\bigskip
\noindent
{\bf Notation.} 
For a vector $x$, $\|x\|_p=(\sum_i |x_i|^p)^{1/p}$ is the standard $\ell_p$-norm. 
For a matrix $A$, let $\vertiii{A}_p = \max_{\lnorm{x}{p} = 1} \lnorm{Ax}{p}, \forall p \in \naturals \cup \{0, \infty\}$ denote the operator norm. 
Then $\maxnorm{A} = \max_i \sum_{j} \abs{A_{ij}}$, $\onorm{A} = \max_j \sum_{i} \abs{A_{ij}}$  and  $\tnorm{A} = \sigma_{\max}(A)$, the maximum singular value. $\ones$ is all ones vector with appropriate dimensions and $\identy$ is identity matrix with appropriate dimensions. $[n] = \{1,2,\ldots,n\}\;, \forall n \geq 1$. For a vector $x \in \reals^n$, $x_i$ ($i \in [n]$) is its $i$-th coordinate.

\section{Our first architecture: RCGAN}
\label{sec:rcgan}

%For simplicity, assume that we know the real prior $P_Y$ on $Y$. 

%In Section \ref{sec:intro} we saw that 
Training a conditional GAN with noisy samples results in a biased generator. 
%To tackle the bias, 
We propose Robust Conditional GAN (RCGAN) architecture which has the following pre-processing, discriminator update, and generator update steps. We assume in this section that the confusions matrix $C$ is known (and the marginal $P_Y$ can easily be inferred), and 
address the case of unknown $C$ in Section \ref{sec:RCGAN-U}. 

\begin{figure}[h]
	\centering
	\includegraphics[width=.35\textwidth]{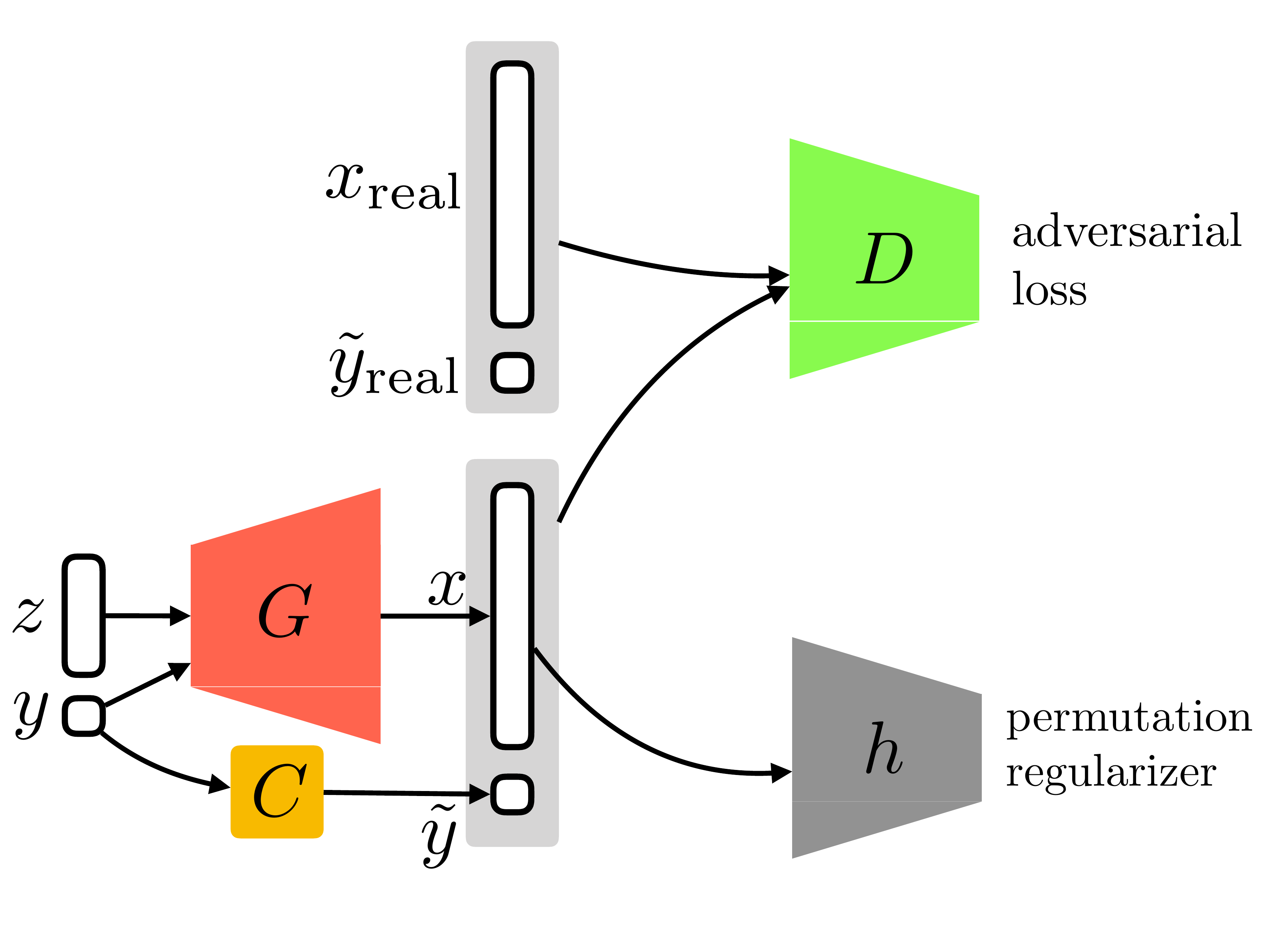}
	\caption{The output $x$ 
	of the conditional generator $G$ is 
	paired with a noisy label $\ty$ corrupted by the channel $C$. 
	The discriminator $D$ estimates   
	whether a given labeled sample is coming from the real data $(x_{\rm real},\tilde{ y}_{\rm real})$ or generated data $(x,\tilde{y})$. 
	The permutation regularizer $h$ is pre-trained on real data. 
	}
	\label{fig:RCGAN}
\end{figure}

\bigskip
\noindent
{\bf Pre-processing:} We train a classifier $h^*$ to predict the noisy label $\ty$ given $x$ under a loss $l$, 
trained on 
$h^* \in \arg\min_{h \in \mathcal{H}} {\mathbb E}_{(x,\ty) \sim \tP_{X,\tY}} [\ell(h(x),\ty)]$, 
where $\mathcal{H}$ is a parametric family of classifiers (typically neural networks) 
and $\tP_{X,\tY}$ is the joint distribution of real $x$ and corresponding real noisy $\ty$.  

\bigskip
\noindent
{\bf D-step:} We train on the following adversarial loss. 
In the second term below, $y$ is generated according to $P_Y$ and corresponding noisy labels are generated by 
corrupting the $y$ according to the conditional distribution $\confus_y$ 
which is the $y$-th row of the confusion matrix (assumed to be known):
\begin{align*}
\max_{D \in \cF  }\; \expectD{(x,\ty) \sim \tP_{X,\tY}}{\phi\left(\D(x,\ty)\right)} + \expectD{\substack{z \sim N, \, y \sim P_{Y} \\ \ty|y \sim \confus_y}}{\phi\left(1- \D(G(z;y), \ty)\right)} \,,
\end{align*}
where $P_Y$ is the true marginal distribution of the labels, $N$ is the distribution of the latent random vector, and 
$\cF$ is a  family of discriminators.

\bigskip
\noindent
{\bf G-step:} We train on the following loss with some  $\lambda > 0$:
\begin{align}
\min_{G \in \mathcal{G}}\; \smallexpectD{\substack{z \sim N, \, y \sim P_{Y} \\ \ty|y \sim \confus_y}}{\phi\left(1- \D(G(z;y), \ty)\right) + {\lambda \, \ell (h^*(G(z;y)), y)} }
\;,
\label{eq:Gstep}
\end{align}
where $\cG$ is a family of generators.  
The idea of using auxiliary classifiers 
have been used to improve the quality of the image and stability of the training, for example in auxiliary classifier GAN (AC-GAN) \cite{OOS16}, 
and improve the quality of clustering in the latent space  \cite{cluster}. 
We propose an auxiliary classifiers $h$, mitigating a {\em permutation error}, which  
we empirically identified on naive implementation of our idea with no regularizers.

\bigskip
\noindent
{\bf Permutation regularizer} (controlled by $\lambda$). 
Permutation error occurs if, when asked to produce samples from a target class, 
the trained generator  produces samples dominantly from a single class but different from the target class. 
We propose a regularizer $h^*$, which predicts the {\em noisy} label $\ty$. 
As long as the confusion matrix is diagonally dominant, 
which is a necessary condition for identifiability, this regularizer encourages the correct permutation of the labels. 
% given the real sample $X$ in the pre-processing step, under a choice of a loss function $\ell:[m]\times[m]\to\reals^+$. 
%In the G-step it penalizes the generator for producing sample $X$ which is hard for the classifier $h^*$ to correctly label as the same samples's label $Y$. 
%If according to the noisy real distribution of $(X,\tY)$, if a point $x$ is more likely to labeled with a particular label $\ty$, this regularizer encourages the generator to produce the same label for the same $x$, thus reducing the permutation error.

%\bigskip 
%\noindent
%{\bf Mixing regularizer} (controlled by $\lambda_2$). Mixing error happens if, when generator is asked to produce samples from one class, it produces a mix of different classes. 
%We propose the mixing regularizer, which tries to minimize the best possible classification loss $\expect{\ell(h_2(x), y)}$ when asked to predict $Y$ given $X$. 
%This encourages that the generated 
%samples from different classes are different. 
%In practice, we only require very weak classifiers like linear classifier for $h_1, h_2$ and standard cross-entropy loss as $\ell$. 
%For the discriminator, we choose {\em projection discriminator} as suggested by our theoretical analyses in the following section. 

\bigskip
\noindent{\bf Theoretical motivation for RCGAN.}
When $\lambda = 0$, we get the standard conditional GAN update steps, albeit one which tries to minimize discriminator loss between the noisy real distribution $\tP$ and the distribution $\tQ$ of the generator when the label is passed through the same  noisy channel parameterized by $\confus$. 
%To theoretically motivate the corruption of the labels of the generated distribution we give the following results. 
The main idea of RCGAN is to 
minimize a certain divergence between noisy real data and noisy generated data. 
For example, the choice of  bounded functions  
$\cF=\{D:\cX\times[m]\to[0,1]\}$ and identity map 
$\phi(a)=a$ leads to a total variation minimization;  
The loss minimized in the 
G-step is the  total variation $d_{\rm TV}(\tP,\tQ)\triangleq \sup_{S\in \cX\times [m]} \{\tP(S) -\tQ(S)\}$ 
between the two distributions with corrupted labels, 
up to some scaling and some shift.  
%and up to the approximation error resulting from having finite samples and not the distribution. 
If we choose 
$\cF=\{D:\cX\times[m]\to [0,1]\}$ and $\phi(a) = \log(a)$, then 
we are minimizing the Jensen-Shannon divergence 
$d_{\rm JS}(\tP,\tQ) \triangleq (1/2) d_{\rm KL}(\tP \| (\tP+\tQ)/2) + (1/2)d_{\rm KL}(\tQ \| (\tP+\tQ)/2)$, 
where $d_{\rm KL}(\cdot \| \cdot)$ denotes the Kullback-Leibler divergence. 
The following theorem 
provides approximation guarantees for 
some common  divergence measures over noisy channel, justifying our proposed practical approach. 
We refer to Appendix \ref{app:prob_dist_up_lb_pf} for a proof. 

%We will first provide multiplicative approximation bounds 
%for population divergence measures, 
%and extend it to additive approximation bounds for 
% sample divergence measures in Section \ref{sec:finite}. 
%a divergence, e.g.~$d_{\rm TV}(P,Q)$, 
%has {\em approximation bounds }

\begin{theorem} \label{thm:prob_dist_up_lb}
	Let $\xy{P}$ and $\xy{Q}$ be two distributions on $\cX \times [\ny]$. Let $\xty{\tP}, \xty{\tQ}$ 
	be the corresponding  distributions when samples from $P, Q$ 
	are passed through the noisy channel given by the confusion matrix $\confus \in \reals^{\ny \times \ny}$ (as defined in Section \ref{sec:noise-model}). 
	If $\confus$ is full-rank, we get,
	\begin{align}
	 \tv{\tP}{\tQ} \;\;\leq  &\;\;\tv{P}{Q}\;\;  \leq \;\; \vertiii{C^{-1}}_\infty \, \tv{\tP}{\tQ} \;,\text{ and } \label{eq:tv_ub_lb} \\
	\frac{1}{8}\, \js{\tP}{\tQ}^2  \;\; \leq  &\;\;\js{P}{Q}\;\; \leq\;\;  \vertiii{C^{-1}}_\infty \sqrt{8\, \js{\tP}{\tQ}}  \;.
	\label{eq:js_ub_lb}
	\end{align}
\end{theorem}
To interpret this theorem,
let $Q$ denote the distribution of the generator. 
% and hence $\tQ$ is the distribution of the generated samples after the label has been corrupted by a noisy channel. 
The theorem implies that when
 the noisy generator distribution $\tQ$ 
 becomes close to the noisy real distribution $\tP$ 
 in total variation or in Jensen-Shannon divergence, 
 then the generator distribution $Q$ must be close to 
 the distribution of real data $P$  in the same metric. 
%Further, if the distance between $P, Q$ is small then distance between $\tP, \tQ$ is also small, which in turn implies that if distance between $P, Q$ can be minimized in the these metrics, it is possible to minimize distance between $\tP, \tQ$. 
This justifies the use of  the proposed architecture RCGAN. 
In practice, we minimize the sample divergence of 
the two distributions, instead of the population divergence as analyzed in the above theorem. 
However, these standard divergences are known to not generalize in training GANs \cite{arora2017generalization}. 
To this end, we provide in Section \ref{sec:theory} analyses on {\em neural network distances}, which are known to generalize, 
and provide finite sample bounds. % including the sample complexity.   
\section{Theoretical Analysis of RCGAN}
\label{sec:theory}

It was shown in \cite{arora2017generalization} that 
standard GAN losses of 
 Jensen-Shannon divergence and Wasserstein distance both 
 fail to generalize with a  finite number of samples. 
% For a precise definition of generalization for training GANs, we refer to \cote{arora2017generalization}. 
% Hence, it was concluded in \cite{arora2017generalization} that 
% small $D_{\rm JS}(P\| Q)$ cannot be achieved, for example, by training a GAN on the empirical cross-entropy loss, 
% which approximates the empirical Jensen-Shannon divergence. 
On the other hand, more recent advances in analyzing GANs in \cite{ZLZ17,BCST18,BMR18} 
show promising generalization bounds 
by either assuming Lipschitz conditions on the generator model or 
by restricting the analysis to certain classes of distributions. 
Under those assumptions, where JS divergence generalizes, 
Theorem \ref{thm:prob_dist_up_lb} 
justifies the use of the proposed RCGAN. 
However, those require  
the distribution to be Gaussian, mixture of Gaussians, or output of a neural network generator, 
for example in \cite{BMR18}.  
%the distribution of real data  is unlikely to satisfy the conditions in , for example.  

In this section, we provide analyses of RCGAN on
a distance that generalizes without any assumptions on the distribution of the real data as proven in \cite{arora2017generalization}:  
{\em neural network distance}. 
Formally, consider a class of real-valued  functions $\cF$ 
 %That is, we cannot use probability distance between the empirical distributions as a surrogate for the probability distance between the actual distributions in a population level. On the other hand, the authors showed that the \NND
%  as defined below in equation \ref{eq:nnd} would generalize well under certain assumptions including parametric class of feasible Discriminators and Lipschitz smoothness. Hence, for the remainder of the paper we will talk only in terms of \NND.
%Let $\cF$ be a class of vector functions such that $\D: \cX \to [0,1]^{\ny}, \; \forall \D \in \cF$, 
 and a function $\phi: [0, 1] \to \reals$ which is either convex or concave. 
 The neural network distance is defined as 
%  between two distributions $P$ and $Q$ on $\cX \times [\ny]$ is defined for convex $\phi(\cdot)$ as,
\begin{align}
\label{eq:nnd}
\nnd{P}{Q} \;\;\; \define \;\;\; \sup_{\D \in \cF} \;\; \expectD{(x,y) \sim P}{\phi\left(\D(x,y)\right)} + \expectD{(x,y) \sim Q}{\phi\left(1 - \D(x,y) \right)} - 
%\phi(0) - \phi(1) \,.
\mu_\phi \,.
\end{align}
where $P$ is the distribution of the real data, $Q$ is that of the generated data, 
and $\mu_\phi$ is the constant correction term to ensure that 
$\nnd{P}{P}=0$. 
We further assume that $\cF$ includes three constant functions 
 $D(x,y)=0$, $D(x,y)=1/2$, and $D(x,y)=1$,
in order to ensure that $\nnd{P}{Q}\geq 0$ and $\nnd{P}{P}=0$, as shown in 
Lemma \ref{lem:metric} in the Appendix. 

The proposed RCGAN with $\lambda = 0$ approximately minimizes the neural network distance 
$\nnd{\tP}{\tQ}$ between the two corrupted distributions. 
In practice, $\cF$ is a parametric family of functions from a specific neural network architecture that the designer has chosen. 
In theory, we aim to identify how the choice of class $\cF$ provides the desired approximation bounds similar to those in Theorem \ref{thm:prob_dist_up_lb}, 
but for neural network distances. 
This analysis leads to the choice of {\em projection discriminator} \cite{miyato2018cgans}  to be used in RCGAN (Remark \ref{rem:proj}).
On the other hand, we show in Theorem~\ref{thm:gen_counter} that  an inappropriate 
choice of the discriminator architecture can cause non-approximation. 
Further, we provide the sample complexity of the approximation bounds in Theorem~\ref{thm:nn_sample}.

We refer to the un-regularized version with $\lambda = 0$ as simply RCGAN. 
In this section, we focus on a class of loss functions called 
Integral Probability Metrics (IPM) where $\phi(x) = x$ \cite{SFG09}. 
This is a popular choice of loss in GANs in practice \cite{STS16,arjovsky2017wasserstein,BSA18} and in analyses \cite{BMR18}. 
We  write the induced neural network distance as $\nndff{\cF}{P}{Q}$, dropping the $\phi$ in the notation.

%
%In this section, we show that 
%the neural network distance 
%
%
%
%
%% as we formally prove in Appendix \ref{lem:metric}
%
%This significantly improves upon 
%
%
%\cite{bora2018ambientgan}
%
%
%
%
%
%NOT divergences, but 
%
%GANs are minimizing neural network distances (As loss in sample dsitance generalizes to population distances). 
%
%
% 
%Ambient GAN minimizing loss of $\nnd {\tP}{\tQ}$
%
%Fundamental question is how is it related to the actual distance $\nnd{P}{Q}$
%
%IT depends on  the confusion matrix $C$, as 
%
%
%
%
%
%It also depends on the the function class $\cF$ that we optimize over. 
%Under a certain condition on $\cF$ (Assumption \ref{eq:mix_inv}), 
%we make this connection precise. 

%%%%%%%%%%%%%%%%%%%%%%%%%%%%%%%%%%%%%%%%%%%%%%%%%%%%%%%%%%%%%%%%%%%%%%

%%%%%%%%%%%%%%%%%%%%%%%%%%%%%%%%%%%%%%%%%%%%%%%%%%%%%%%%%%%%%%%%%%%%%%
\subsection{Approximation bounds for neural network distances}

%\begin{enumerate}
%	\item concave $\phi(\cdot)$ as,
%	\begin{align}
%	\nnd{P}{Q} \define \sup_{\D \in \cF} \;\; \expectD{(x,y) \sim P}{\phi\left(\D(x)_y\right)} + \expectD{(x,y) \sim Q}{\phi\left(1 - \D(x)_y\right)} - 2\phi(1/2)\,, \label{eq:nnd}
%	\end{align}	
%	\item convex $\phi(\cdot)$ as
%	\begin{align}
%	\label{eq:nnd}
%	\nnd{P}{Q} \define \sup_{\D \in \cF} \;\; \expectD{(x,y) \sim P}{\phi\left(\D(x)_y\right)} + \expectD{(x,y) \sim Q}{\phi\left(1 - \D(x)_y\right)} - \phi(0) - \phi(1)\,.
%	\end{align}
%\end{enumerate}

 We define an operation $\vopsymb$ over  
 a matrix $T \in \reals^{\ny \times \ny}$ and 
 a class $\cF$ of functions on $\cX\times [\ny] \to \reals$ as 
\begin{align}
	\label{eq:vop}
		\vop{T}{\cF} \;\; \triangleq \;\;  \Big\{  f  \in \cF \;|\; f(x,y)  = \sum_{\ty\in[\ny]} T_{y\ty}\, f(x,\ty)    \Big\}\,.
\end{align}
This makes it convenient to represent the neural network distance 
corrupted by noise with a confusion matrix $C\in\reals^{m\times m}$, 
where $C_{y\ty}$ is the probability a label $y$ is corrupted as $\ty$. 
Formally,  it follows from \eqref{eq:nnd} and \eqref{eq:vop} that $d_{\cF}(\tP,\tQ) = d_{\vop{C}{\cF}}(P,Q)$. 
We refer to Appendix \ref{app:gen_ub_lb_pf} for a  proof. 
For $\nndff{\cF}{\tP}{\tQ}$ to be a good approximation of $\nndff{\cF}{P}{Q}$, 
we show that the following condition is  sufficient. 

\begin{assume} We assume that the class of discriminator functions $\cF$ can be decomposed into three parts  
%\begin{align}
$\cF = \{ f_1 + f_2 + c \,|\, f_1 \in \cF_1, f_2 \in  \cF_2\} $ such that $c\in\reals$ is any constant and 
	\begin{itemize}
		\item $\cF_1$ satisfies the {\em inclusion condition}: 
		 	\begin{eqnarray}
				 \vop{T}{\cF_1} \;\; \subseteq \;\; \cF_1\;, 
				 \label{eq:def_inclusion}
			\end{eqnarray}
			for all  $\maxnorm{T} \triangleq  \max_i \sum_{j} \abs{T_{ij}} = 1$; and 
		\item $\cF_2$ satisfies the  {\em label invariance condition}: 	 	there exists a class $\cF^{(x)}$ of sets of functions,
			parametrized by $x\in \cX$, such that  
			\begin{eqnarray} 
				\cF_2 \;\; =  \;\; \big\{\, \alpha f(x,y) \,\,|\,\, f(x,y) = f(x), f(x) \in \cF^{(x)}  , \alpha\in[0,1] \, \big\} \;.
				\label{eq:def_label}
			\end{eqnarray} 
	\end{itemize}
%\end{align}
\label{eq:mix_inv}
\end{assume}
We discuss the necessity and practical implications of this assumption in Section \ref{sec:assume}, 
and give examples satisfying these assumptions in Remarks \ref{rem:proj} and 
\ref{rem:proj_example}. 
%Appendix~\ref{sec:example}.
%\ref{rem:proj_example} and \ref{rem:proj}. 
Notice that a trivial class with a single 
constant zero function satisfies both inclusion and label invariance conditions. 
For example, we can choose $c=0$ and also choose to set either $\cF_1= \{f(x,y)=0\}$ or 
$\cF_2= \{f(x,y)=0\}$, in which case $\cF$ only needs to satisfy either one of the conditions in Assumption \ref{eq:mix_inv}. 
The flexibility that we gain by allowing the set addition $\cF_1+\cF_2$ is critical in applying these conditions to practical discriminators, 
especially in proving Remark \ref{rem:proj}. 
Note that in the inclusion condition in Eq.~\ref{eq:def_inclusion}, 
we require the condition to hold for all max-norm bounded set: $\{T:   \max_i \sum_{j} \abs{T_{ij}} = 1\}$. 
The reason a weaker condition of all row-stochastic matrices, $\{T: \sum_j T_{ij} = 1\}$, does not suffice is that 
in order to prove the upper bound in Eq.~\ref{eq:main1}, 
we need to apply the invariance condition to $\vertiii{C^{-1}}_\infty^{-1} C^{-1} \circ \cF$. 
This matrix $\vertiii{C^{-1}}_\infty^{-1} C^{-1}$ is not row-stochastic, but 
still max-norm bounded.

We first show that Assumption \ref{eq:mix_inv} is sufficient for approximability of the neural network distance from corrupted samples.  
	For two distributions 
	$\xy{P}$ and $\xy{Q}$ on $\cX \times [\ny]$,  
	let $\xty{\tP}$ and  $\xty{\tQ}$ be the corresponding 
	corrupted distributions respectively, 
	where the label $Y$ is passed through the noisy channel defined by the confusion matrix $\confus \in \reals^{\ny \times \ny}$,  
	i.e.~$\tP(x,\ty) = \sum_y P(x,y) C_{y,\ty}$. 
	%noisy-labeled distributions when samples from $P, Q$ are passed through the noisy-channel given by the confusion matrix $\confus \in \reals^{\ny \times \ny}$ . 
	
\begin{theorem} \label{thm:gen_ub_lb}
	%Let $P, Q, C$ satisfy the same assumptions as in Theorem \ref{thm:prob_dist_up_lb}. 
	If a class of functions $\cF$ satisfies Assumption \ref{eq:mix_inv}, then 
	\begin{align}
		 \nndff{\cF}{\tP}{\tQ} \;\; \leq \;\; \nndff{\cF}{P}{Q} \;\; \leq \;\; \vertiii{C^{-1}}_\infty  \nndff{\cF}{\tP}{\tQ} \;,
	\label{eq:main1}
	\end{align}
	where %$\vertiii{C^{-1}}_\infty = \max_i \sum_j |(C^{-1})_{ij}|$ is always at least one, and
	we follow the convention that $\vertiii{C^{-1}}_\infty =\infty $ if $C$ is not full rank. 
\end{theorem}

We refer to Appendix  \ref{app:gen_ub_lb_pf} for a proof. 
%Let $\vcF$ be a class of vector valued functions such that $f: \cX \to \reals^\ny ,\; \forall f \in \vcF$ and $0\, \ones \in \vcF$. In this section, we assume that the measure function $\phi(x) = x$. For ease of notation we drop the $\phi$ to define the \NND when measuring function is $\phi(x) = x$ as,
% \begin{align}
% \nndff{\vcF}{P}{Q} &\define \nndfp{\vcF}{x}{P}{Q} \nonumber \\
% &= \sup_{D \in \scF} \;\; \expectD{(x,y) \sim P}{D(x,y)} + \expectD{(x,y) \sim Q}{1 - D(x,y)} - 1 \nonumber \\
% &= \sup_{D \in \scF} \;\; \expectD{(x,y) \sim P}{D(x,y)} - \expectD{(x,y) \sim Q}{D(x,y)}
% \end{align}
%
%\begin{theorem} \label{thm:gen_ub_lb}
%	Let $P, Q, C$ satisfy the same assumptions as in Theorem \ref{thm:prob_dist_up_lb}. Then if class of vector functions, $\vcF$ satisfies,
%	\begin{align}
%		\vop{T}{\vcF} \subseteq \vcF \text{ for all } \maxnorm{T} = 1 &&\text{\it(projection invariance)}, \label{eq:mix_inv}
%	\end{align}
%	where $\vop{T}{\vcF}$ as defined in \eqref{eq:vop} and if $\confus$ is full-rank then,
%	\begin{align}
%	\normconfus \nndff{\vcF}{P}{Q} \leq \nndff{\vcF}{\tP}{\tQ} \leq \nndff{\vcF}{P}{Q}
%	\end{align}
%\begin{proof}
%Proof is in Section  \ref{app:gen_ub_lb_pf}.
%\end{proof}
%\end{theorem}
This gives a sharp characterization on how two distances are related: 
the one we can minimize in training RCGAN (i.e.~$\nndff{\cF}{\tP}{\tQ}$) 
and the true measure of closeness (i.e.~$\nndff{\cF}{P}{Q}$). 
Although the latter cannot be directly evaluated or minimized, 
RCGAN is approximately minimizing the true neural network distance $\nndff{\cF}{P}{Q}$ as desired. 
%Together with well-known results that 
%neural network distances are \cite{arora2017generalization} ???? . 

The lower bound proves   a special case of the data-processing inequality. 
Two random variables from $P$ and $Q$ 
get closer in neural network distance, when passed through a stochastic transformation. %a label corruption process defined by the confusion matrix $C$. 
The upper bound puts a limit on  how much closer $\tP$ and $\tQ$ can get, depending on the noise level. 
This fundamental trade-off  is 
captured by $\vertiii{C^{-1}}_\infty$. 
Under the noiseless case where $C$ is the identity matrix, we have $\vertiii{C^{-1}}_\infty=1$ and 
we recover a trivial fact that the two distances are equal. 
On the other extreme, if $C$ is rank deficient, we use the convention that 
 $\vertiii{C^{-1}}_\infty=\infty$ and the two distances can be arbitrarily different. 
%For any regime in between, the above theorem gives a sharp characterization. 
The approximation factor of $\vertiii{C^{-1}}_\infty$ 
captures  how much the space $\cF$  can shrink by the noise $C$. 
This coincides with Theorem \ref{thm:prob_dist_up_lb}, where a similar trade-off was 
identified for the TV distance.
Next remark  shows
that these bounds cannot be tightened 
for general $P$, $Q$, and $\cF$.
A proof is provided in Appendix~\ref{sec:tight}.

\begin{remark} 
	\label{rem:up_lb_tight}
	For any full-rank confusion matrix $C \in \reals^{d_1 \times d_2}$, 
	 there exist pairs of distributions $(P_1,Q_1)$ and $(P_2,Q_2)$, and 
	 a function class $\cF$ satisfying Assumption \ref{eq:mix_inv}, 
	  such that
	\begin{itemize}
		\item[1.] $\nndff{\cF}{\tP_1}{\tQ_1} \;\; = \;\; \nndff{\cF}{P_1}{Q_1}$, and 
		\item[2.] $\nndff{\cF}{P_2}{Q_2} \;\; = \;\; \vertiii{C^{-1}}_\infty  \nndff{\cF}{\tP_2}{\tQ_2}$\;.
	\end{itemize}
\end{remark}
%in the following remark we show that both the upper and lower bounds are tight for Theorem \ref{thm:gen_ub_lb} and the Total Variation in Theorem \ref{thm:prob_dist_up_lb}. 

%Further, the approximation guarantee is determined by the $\ell_\infty$ norm of the inverse of $C$, which gives a sharp characterization.  
Theorem \ref{thm:gen_ub_lb} shows that $(i)$ RCGAN can learn the true conditional distribution, justifying its use; and 
$(ii)$ performance of RCGAN is determined by how noisy the samples are via $\vertiii{C^{-1}}_\infty$. 
There are still two loose ends. 
First, does practical implementation of RCGAN architecture satisfy the inclusion and/or label invariance assumptions? 
Secondly, in practice we cannot minimize 
$\nndff{\cF}{\tP}{\tQ}$ as we only have a finite number of samples. 
How much do we lose in this finite sample regime? 
We give precise answers to each question in the following two sections.

%%%%%%%%%%%%%%%%%%%%%%%%%%%%%%%%%%%%%%%%%%%%%%%%%%%%%%%%%%%%%%%%%%%%%%
% 
%\textcolor{red}{
%but first we give examples to help interpret the approximation factor $\|C^{-1}\|_\infty$. 
%%
%\begin{corollary}
%	Under the assumptions of Theorem \ref{thm:gen_ub_lb} and when $Y$ is a binary random variable and the confusion matrix is,
%	\begin{align}
%	\confus = \begin{bmatrix}
%	1-\rho_1 & \rho_1 \\
%	\rho_2	& 1-\rho_2
%	\end{bmatrix}\,,
%	\end{align}
%	we have $\frac{1-\rho_1 - \rho_2}{1 + \abs{\rho_1 - \rho_2}} \nndff{\cF}{P}{Q} \leq \nndff{\cF}{\tP}{\tQ} \leq \nndff{\cF}{P}{Q}$
%\end{corollary}
%}

%%%%%%%%%%%%%%%%%%%%%%%%%%%%%%%%%%%%%%%%%%%%%%%%%%%%%%%%%%%%%%%%%%%%%%
\subsection{Inclusion and label invariance assumptions}
\label{sec:assume}

% implies that RCGAN algorithm can learn the true conditional distribution in the \NND sense if the class of discriminator functions satisfy the invariance property \ref{eq:mix_inv}. That is, if the we minimize the \NND between noisy real distribution $\tP$ and noisy fake distribution $\tQ$, then the \NND between real distribution $P$ and the fake distribution $Q$ generated by the conditional Generator is also minimized. The above Theorem \ref{thm:gen_ub_lb} directly gives the following corollary.

%Next, we will prove that the sufficient condition in Theorem \ref{thm:gen_ub_lb} is not vacuous.

% many choices of conditional GANs. 

Several class of functions satisfy  Assumption \ref{eq:mix_inv} (c.f.~Remark \ref{rem:proj_example}).   
For RCGAN, we propose 
a popular state-of-the-art discriminator for conditional GANs known as the {\em projection discriminator} \cite{miyato2018cgans}, 
parametrized by $V \in \reals^{m \times d_V}$, $v\in\reals^{d_v}$, and $\theta\in\reals^{d_\theta}$: 
\begin{align}
\D_{V,v,\theta}(x,y) \;\; = \;\;  {\rm vec}(y)^T \, V\,  \psi(x;\theta)  \,+\,  v^T\,\psi'(x;\theta)\;, \label{eq:project}
\end{align}
where $\psi(x;\theta) \in \reals^{d_V}$ and $\psi'(x;\theta)\in\reals^{d_v}$ are vector valued parametric functions for some integers $d_V,\,d_v$, 
and ${\rm vec}(y)^T = [{\mathbb I}_{y=1} , \ldots, {\mathbb I}_{y=m} ]$. 
The first  term satisfies the inclusion condition, 
 as any operation with $T$ can be absorbed into $V$. 
The second term is label invariant as it does not depend on $y$. 
This is made precise in the following remark, whose proof is provided in Appendix \ref{app:proj_example_pf}.
Together with this remark, the approximability result in Theorem \ref{thm:gen_ub_lb}  justifies the use of 
projection discriminators in RCGAN, which we use in all our experiments. 
\begin{remark} 
	The class of projection discriminators $\{\D_{V,v,\theta}(x,y)\}_{V\in\cV_1,v\in\cV_2 ,\theta\in\Theta}$ defined in Eq.~\ref{eq:project} satisfies Assumption  \ref{eq:mix_inv} for any $\psi$,  $\psi'$, and $\Theta$, if 
$
		\cV_1 = \big\{ \, V\in\reals^{m\times d_V} \,\big|\, \max_{i} |V_{ij}| \leq 1 \text{ for all } j\in[d_V]  \, \big\}  \;,
		\text{ and }
		\cV_2 = \big\{ \, v \in \reals^{d_v} \,\big|\, \|v\| \leq 1 \, \big\}\;.
		 %\alpha v\in \cV_2,\;\;\text{ for all }\;\; v \in \cV_2,\;\; \text{ and }\;\; \alpha\in[0,1]\;.
$
	\label{rem:proj}
\end{remark}
Other choices of $\cV_1$ and $\cV_2$ are also possible. 
For example, $\cV_1'=\{V \in\reals^{m\times d_V} | \sum_j \max_i |V_{ij}| \leq 1 \}$
or $\cV_1''=\{V \in\reals^{m\times d_V} | \maxnorm{V} = \max_i \sum_j |V_{ij}| \leq 1 \}$
are also sufficient. 
 We find the proposed choice of $\cV_1$ easy to implement, 
 as a column-wise $L_\infty$-norm normalization via projected gradient descent.  
We describe implementation details in Appendix \ref{app:implementation}.

%In Appendix \ref{sec:nec}, we show that Assumption \ref{eq:mix_inv} is also necessary. 

% -------------------------------------------------------------------------------------------------------------------------------------------------------------------------
%\section{Necessity of Assumption \ref{eq:mix_inv} } 
%\label{sec:nec}

Next, we ask if Assumption \ref{eq:mix_inv} is necessary also. 
We show that for all pairs of distributions $(P,Q)$ satisfying the following technical conditions, and all confusion matrix $C$, 
there exists a  class $\cF$ where approximation bounds  in \eqref{eq:main1} fail. 
%Ideally, we would want to show that 
%for all $\cF$ that violates the condition, there exist $C,P$, and $Q$, 
%such that the bounds in \eqref{eq:main1} does not hold. 
%This requires constructing 
%counter examples for every function class $\cF$, which is outside the scope of this paper. 
%Instead, we prove a slightly weaker statement in the following. 

\begin{assume} We consider a pair of distributions $P_{X,Y}$ and $Q_{X,Y}$ and a confusion matrix $C$ satisfying the following conditions:  
\begin{itemize}
	\item The random variable $X$ conditioned on $Y=y$ is a 
	continuous random variable with density functions  
	$dP_{X|Y=y}$ and $dQ_{X|Y=y}$, respectively. %$P_X,\; Q_X$ are continuous distributions in $\cX$ with valid probability densities $p(x), q(y)$; and 
 	\item There exists $\, S \subseteq \cX \text{ such that } P_X(S)+Q_X(S) > 0$,  and  
	$P_{X,Y}(x,\cdot) - Q_{X,Y}(x,\cdot)$  is not a right eigenvector of  $C$, for all $ x \in S$, 
	where $P_{X,Y}(x,\cdot) = [P_{X,Y}(x,1 ) , \cdots, P_{X,Y}(x,m)]^T$. 
\end{itemize}

\label{assume2}
\end{assume}

A pair $(P,Q)$ violating the above assumptions   
either has $X$ that is a mixture of continuous and discrete distribution, or 
all $(P(x,\cdot) - Q(x,\cdot))$'s are aligned with the right 
eigenvectors of $C$. 

\begin{theorem} \label{thm:gen_counter}
	For all sufficiently small $\epsilon > 0$, all distributions $P_{X, Y}$ and  $Q_{X, Y}$ 
	satisfying Assumption \ref{assume2}, 
	and all full-rank 
	$\confus \in \reals^{\ny \times \ny}$, there exist 
	$\cF_3$  not satisfying Assumption  \ref{eq:mix_inv}, such that 
	\begin{align}
	&\nndff{\cF_3}{\tP}{\tQ} \leq O_\epsilon(\epsilon) \;\; \text{ and } \;\; \nndff{\cF_3}{P}{Q} \geq O_\epsilon(1), 
	\end{align}
	and $\cF_4$ not satisfying Assumption  \ref{eq:mix_inv}, such that 
	\begin{align}
	& \nndff{\cF_4}{\tP}{\tQ} \geq O_\epsilon(1) \;\; \text{ and } \;\; \nndff{\cF_4}{P}{Q} \leq O_\epsilon(\epsilon) \,.
	\end{align}
\end{theorem}
We refer to Appendix \ref{app:gen_counter_pf} for a proof. 
This implies that 
some assumptions 
on the function class $\cF$ 
are necessary, such as those in Assumption  \ref{eq:mix_inv}.
%we need to restrict 
Without any restrictions,  
we can find bad examples where the two distances  
$\nndff{\cF}{P}{Q}$ and $\nndff{\cF}{\tP}{\tQ}$ are arbitrarily different for any $C$, $P_{X,Y}$, and $Q_{X,Y}$. 

% In all our experiments, we use this projection discriminator. 

% The above theorem proves that for almost all real distrbution $P$ and generator distribution $Q$ there exists class of discriminators which does not satisfy projection invariance \ref{eq:mix_inv} and hence RCGAN would fail to learn the correct conditional distribution.

%%%%%%%%%%%%%%%%%%%%%%%%%%%%%%%%%%%%%%%%%%%%%%%%%%%%%%%%%%%%%%%%%%%%%%
\subsection{Finite sample analysis} 
\label{sec:finite} 

In practice, we do not have access to the probability distributions $\tP$ and $\tQ$. 
Instead, we observe a set of samples of a finite size $n$, from each of them. 
In training GAN, we minimize the 
{\em  empirical}  neural network distance, $\nndff{\cF}{\htP}{\htQ}$, 
where   $\htP$ and $\htQ$ denote the  empirical distribution of $n$ samples. 
Inspired from the recent generalization results in \cite{arora2017generalization}, 
we show that this empirical distance minimization leads to 
small $d_{\cF}(P,Q)$ up to  an additive error that vanishes with an increasing sample size.  
%More specifically, we give an upper bound on the number of noisy samples required to learn the true conditional generator using the RCGAN algorithm.
As shown in \cite{arora2017generalization}, 
Lipschitz and bounded function classes are critical in 
achieving sample efficiency for GANs.  
We follow the same approach over  a similar function class. 
Let 
	\begin{eqnarray}
	\cF_{p,L} \;\; =\;\;  \{D_u(x,y) \in [0,1] \,|\, \text{ $D_u(x,y)$ is $L$-Lipschitz in $u$ and } u \in \cU \subseteq \reals^p\}\;,  
	\label{eq:def_lipschitz}
	\end{eqnarray}
be a class of bounded functions with parameter $u\in\reals^p$. 
We say that $\cF$ is $L$-Lipschitz in $u$ if 
\begin{align}
	\abs{D_{u_1}(x, y) - D_{u_2}(x, y)} \; \leq \; L \|u_1 - u_2\| \;,\;\;\;\; \forall u_1, u_2 \in \cU,\; x \in \cX,\; y \in [\ny].
 	\label{eq:param_lip}
\end{align}
\begin{theorem}  
	\label{thm:nn_sample}
%	Let $\cF = \{D_u(x,y) \in [0,1] \,|\, u \in \cU\}$ be a function class parameterized by $u$ for some  $\cU \subseteq \reals^p$, $p \in \mathbb{N}$ , and $\cF$ be $L$-Lipschitz in the parameter as defined in \eqref{eq:param_lip} and 
For any class $\cF_{p,L}$ of 
bounded Lipschitz functions $D_u(x,y)$ 
satisfying Assumption $\ref{eq:mix_inv}$, 
% $\cF_{p,L} - 1/2$ satisfy the inclusion condition 
%	defined in \eqref{eq:def_inclusion}  of Assumption $\ref{eq:mix_inv}$ ??? (why not just assumption 1?). T
there exists a universal constant $c > 0$ such that 
	\begin{align}
%		\abs{\nndff{\cF}{P}{Q} - \nndff{\cF}{\htP}{\htQ}} \;\; \leq \;\; \epsilon\,\normconfusinv + (\normconfusinv - 1)\,\nndff{\cF}{\htP}{\htQ} \,, 
		\nndff{\cF_{p,L}}{\htP}{\htQ} - \epsilon  \;\; \leq \;\; \nndff{\cF_{p,L}}{P}{Q}   \;\; \leq \;\; \normconfusinv \, \big( \nndff{\cF_{p,L}}{\htP}{\htQ} + \epsilon \big)\,, 
		\label{eq:generalize}
	\end{align}
	with probability at least $1 - e^{-p}$ 
	for any $\varepsilon>0$ and $n$ large enough, $
n \;\; \geq \;\; ({c  \,p\,}/{\epsilon^{2}})\,\log\left({ p L }/{\epsilon } \right)\;.$
\end{theorem} 
%\textcolor{red}{Kiran: check the changes I made, and update the proof accordingly.}
We refer to Appendix \ref{app:nn_sample_pf} for a proof. 
%From equation \eqref{eq:generalize}, we can see that if GAN training achieves a small  the noisy empirical distance $\nndff{\cF}{\htP}{\htQ}$, 
%the actual distance goes to $O(\epsilon)=O(1/\sqrt{n})$, thereby recovering a generator distribution $Q$ close to the real distribution $P$. 
% For the noiseless setting with $\normconfusinv = 1$, this recovers the generalization bound in \cite{arora2017generalization}.
This justifies the use of the proposed RCGAN which minimizes $d_\cF(\tP_n , \tQ_n)$, 
as it leads to the generator $Q$ being close to the real distribution $P$ in neural network distance, $d_\cF(P,Q)$. 
These bounds inherit the approximability of the population version in Theorem \ref{thm:gen_ub_lb}. 
% where we identified the dependence in $C$ captured via its key quantity $\vertiii{C^{-1}}_\infty$.

%%%%%%%%%%%%%%%%%%%%%%%%%%%%%%%%%%%%%%%%%%%%%%%%%%%%%%%%%%%%%%%%
\section{Our second architecture: RCGAN-U} 
\label{sec:algorithms}
\label{sec:RCGAN-U}

%In Section \ref{sec:exp}, we show that RCGAN works well when the noise model, $\confus$ is known. 
%However, 
In many real world scenarios the confusion matrix $C$ is unknown.  
We propose RCGAN-Unknown (RCGAN-U) algorithm which jointly estimates the real distribution $P$ and the  noise model $C$. 
The pre-processing and D steps of the RCGAN-U are the same as those of RCGAN, assuming the current guess $M$ of the confusion matrix. 
As the G-step in \eqref{eq:Gstep} is not differentiable in $C$, we use the following reparameterized estimator of the loss, motivated by 
similar technique in training classifiers from noisy labels:  
%The G-step is modified as: 
\begin{align*}
\min_{G \in \mathcal{G}, M \in \mathcal{C}}\; \smallexpectD{\substack{z \sim N \, y \sim P_{Y}}}{\phi_{M}\left(G(z;y), y, \D\right) + {\lambda\,l(h^*(G(z;y)), y)} }\,
\end{align*}
where $\mathcal{C}$ is the set of all  transition matrices and $\phi_M(x,y,D) = \sum_{\ty \in [\ny]} M_{y \ty} \, \phi(1-D(x,\ty))$.

\section{Experiments}
\label{sec:exp}

%{\color{red}
% We make our implementation of RCGAN and RCGAN-U available online at 
% \textcolor{blue}{Github ???}, with all the details to reproduce our experiments. } 
Implementation details are explained in Appendix \ref{app:implementation}. 
We consider one-coin based models, which are parameterized by their label accuracy probability $\alpha$. In this model a sample with true label $y$ is flipped uniformly at random to label $\ty$ in $[\ny] \setminus \{y\}$ with probability $1-\alpha$. The entries of its confusion matrix $\confus$, will then be $\confus_{ii} = \alpha$ and $\confus_{i\neq j} =(1- \alpha)/(\ny-1)$, where $\ny$ is the number of classes. We call this model {\em{uniform flipping}} model.
We train proposed GANs on MNIST and CIFAR-$10$ datasets \cite{lecun1998mnist,krizhevsky2009learning} and compare them to two baselines. Code to reproduce our experiments is available at \url{https://github.com/POLane16/Robust-Conditional-GAN}.

\bigskip
\noindent
{\bf Baselines.} First is the {\em biased GAN}, which is a conditional GAN applied directly on the noisy data. The loss is hence biased, and the true conditional distribution is not the optimal solution of this biased loss. 
Next natural baseline is using de-biased classifier as the discriminator, 
motivated by the approach of \cite{natarajan2013learning} on learning classifiers 
from noisy labels.   
%de-bias the discriminator loss using the techniques developed in \cite{natarajan2013learning}. 
%corresponding to the real labels.  
The main insight is to  modify the loss function according to  $C$, 
such that  in expectation the loss matches that of the clean data. 
We refer to this approach as {\em unbiased GAN}. 
Concretely, when training the discriminator, 
we propose the following (modified) de-biased loss:
\begin{align}
\max_{D \in \cF}\; {\mathbb E}_{(x,\ty) \sim \tP_{X,\tY}}\big[{
\sum_{y\in[m]} 
(C^{-1})_{\ty y} \phi\left(\D(x,y)\right)}\big] + {\mathbb E}_{\substack{z \sim N \, y \sim P_{Y} }} \big[{\phi\left(1- \D(G(z;y), y)\right)  } \big]\,.\label{eq:debiased_loss} 
\end{align} 
This  is unbiased, as the first term 
is equivalent to $\E_{(x,y)\sim P_{X,Y}} [\phi(D(x,y))] $, which  is the standard GAN loss with clean samples.   
%Note that unbiased GAN critically relies on the knowledge of the confusion matrix $C$. 
However, such de-biasing  is sensitive to the condition number of $C$,  
and can become numerically unstable for noisy channels 
as $C^{-1}$ has large entries \cite{khetan2017learning}.  
For both the dataset, we use linear classifiers for permutation regularizer of the RCGAN-U architecture.

\begin{figure}[h]
\begin{center}
\includegraphics[width=0.4\textwidth]{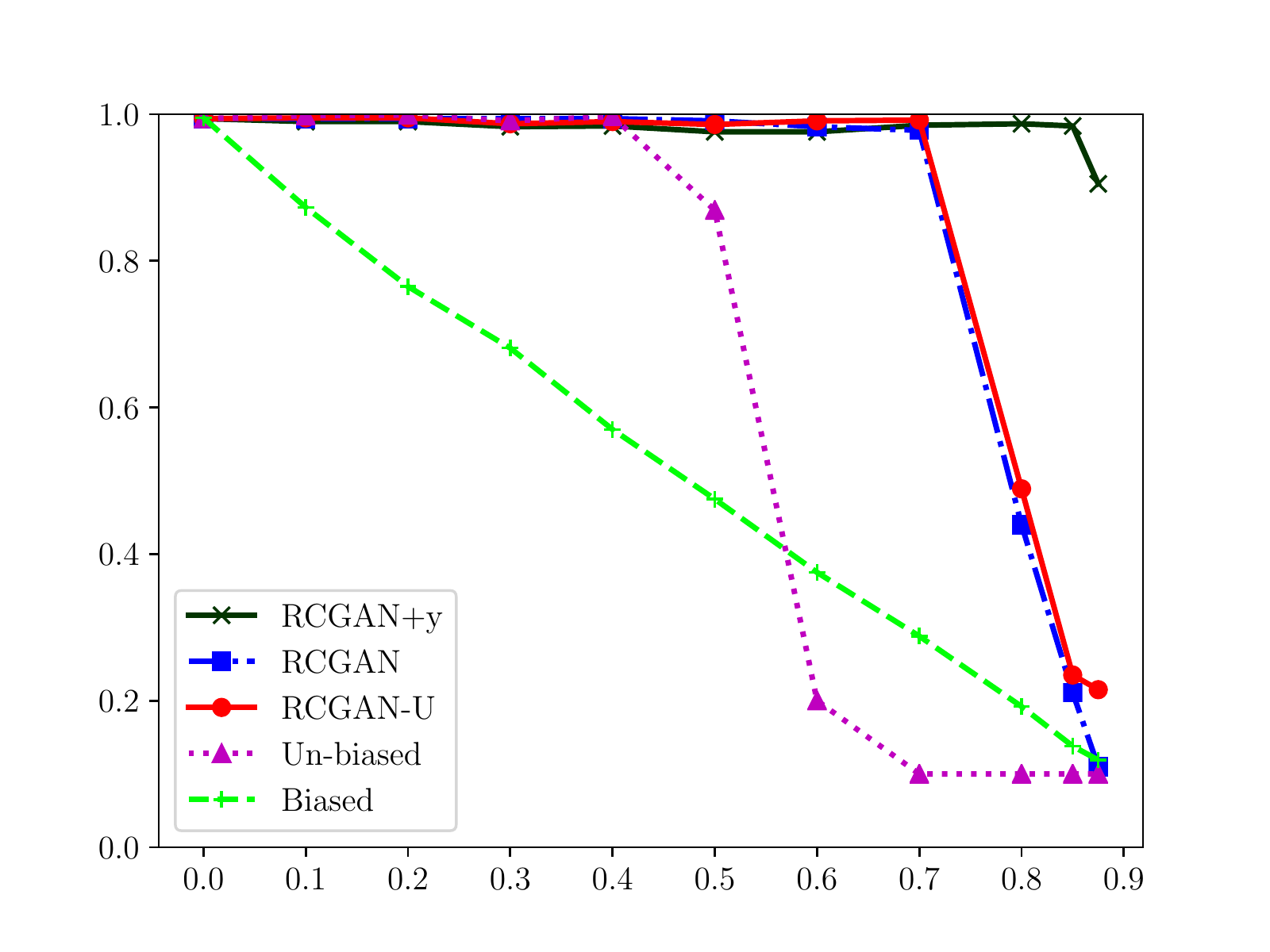}	
\put(-151,-5){\small{ noise in the real data ($1-\alpha$)}}	
\put(-165,130){{generator label accuracy}}
%\put(-49,23){\tiny{$(\lambda$$=$$0)$}}
\includegraphics[width=0.4\textwidth]{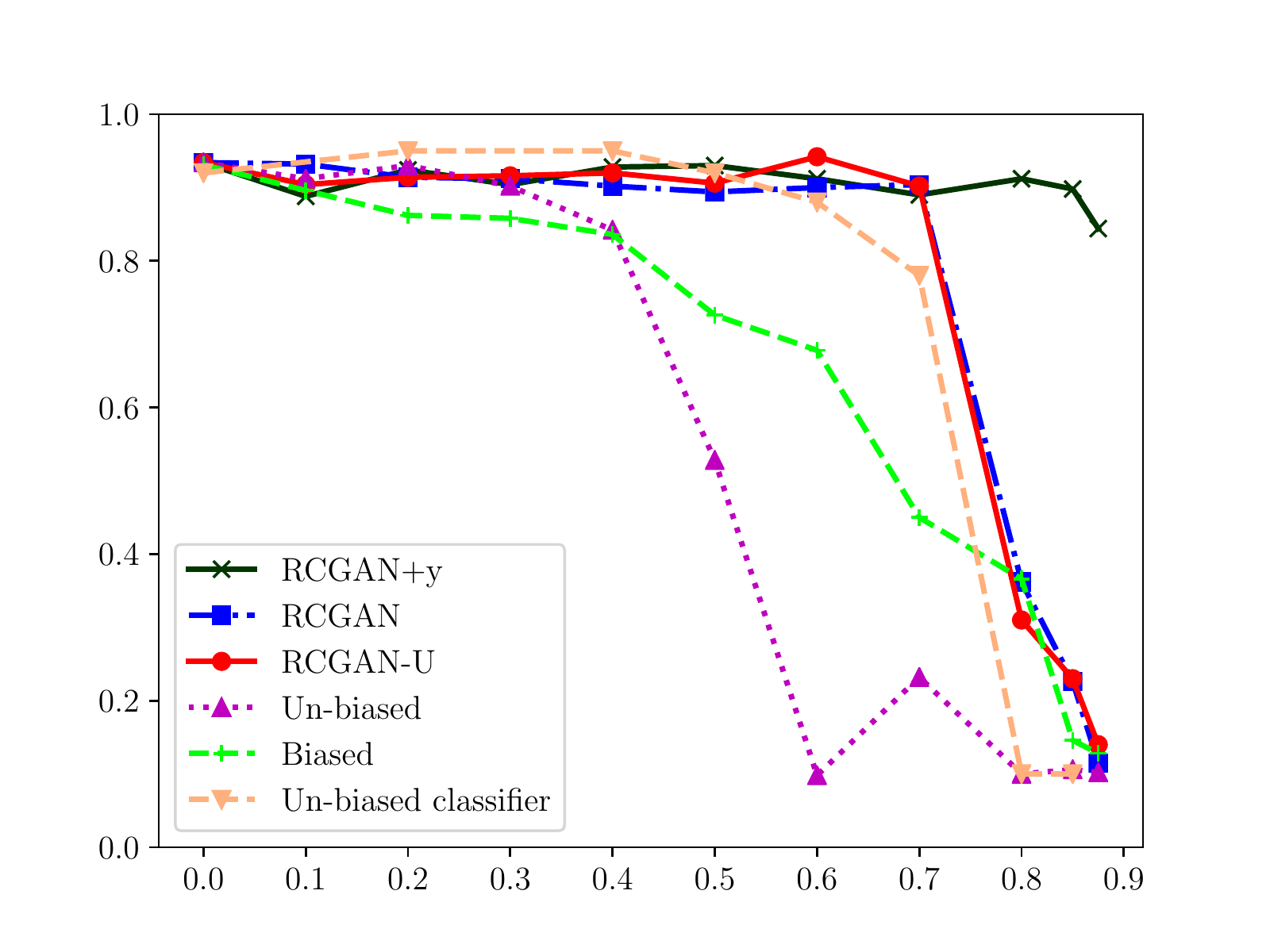}	
\put(-151,-5){\small{ noise in the real data ($1-\alpha$)}}	
\put(-165,130){{label recovery accuracy}}
%\put(-49,28.5){\tiny{$(\lambda$$=$$0)$}}
\caption{
	 Noisy MNIST dataset: Our RCGAN models consistently improves upon all competing baseline approaches in generator label accuracy (left). The trend continues in label recovery accuracy (right), where our proposed RCGAN-classifiers  improves upon {\em unbiased classifier} \cite{natarajan2013learning}, which is one of the state-of-the-art approaches  tailored for  label recovery. 
	 The numerical values of the data points are given in a table in Appendix~\ref{app:A}.
	 %Surprisingly RCGAN-U, without knowledge of true noise model $C$ performs as good as RCGAN which knows $C$.
%	The left panel shows label accuracy of the RCGAN-U generator trained on noisy MNIST dataset, where class labels are flipped uniformly at random with probability $1-\text{\rm{`label accuracy of the real data'}}$ .The right panel shows accuracy in recovering the true real labels of the noisily-labeled real MNIST images using the RCGAN-U-Classifier.}
	}
\label{fig:mnist1}
\end{center}
\end{figure}

\subsection{MNIST}\label{sec:mnist_expt}
We train five architectures on MNIST dataset corrupted by the uniform flipping noise: 
RCGAN+y, RCGAN, RCGAN-U, unbiased GAN, and biased GAN. RCGAN+y architecture has the same architecture as RCGAN but the input to the first layer of its discriminator is concatenated with a one-hot representation of the label. We discuss our techniques to overcome the  challenges involved in training RCGAN+y in Appendix \ref{app:implementation}. 
%We run experiments for $\alpha$ ($1.0$-noise) in the range $[0.125, 1.0]$.

Conditional generators can be used to generate samples $x$ from a particular class $y$, in the classes it learned. We then can use a pre-trained classifier ${f}$ to compare $y$ to the true class of the sample, ${f}(x)$ (as perceived by the classifier ${f}$). 
We compare the {\em{generator label accuracy}} defined as  ${\mathbb E}_{y\sim P_Y, Z\sim N}[{\mathbb I}_{\{y=f(G(z,y))\}}]$, 
%as the accuracy of the label $y$ of $x$ according to the conditional generator when compared to the ``true'' label, ${f}(x)$ as estimated by the pre-trained classifier. The more accurate a conditional generator is, the higher will be the its generator label accuracy. Thus a
%A  robust  GAN achieve higher accuracy even when trained on noisy labels. 
in Figure \ref{fig:mnist1}, left panel. %we show the generator label accuracy of the conditional generator for the five methods. 
We generated $10$k labels chosen uniformly at random and corresponding conditional samples from the generators, and calculated the generator label accuracy using a CNN classifier pre-trained on the clean MNIST data to an accuracy of 99.2\%. The proposed RCGAN significantly improves upon the competing baselines, and achieves almost perfect label accuracy until a high noise of $\alpha=0.3$. RCGAN+y further improves upon RCGAN and to gain very high accuracy even at $\alpha=0.125$. %In comparison, the best unbiased GAN we could train starts to fail at a significantly smaller noise around $\alpha=0.5$. 
The high accuracy of RCGAN-U suggests that robust training is possible 
without prior knowledge of the confusion matrix $C$. 
As expected, biased GAN has an accuracy of approximately $1-\alpha$.

An immediate application of robust GANs is 
recovering the true labels of the noisy training data, 
which  is an important and challenging problem in crowdsourcing. 
%We are provided with $n$ samples from a distribution $P$ and a single noisy-label for each example and the goal is to recover true labels of these examples. 
We propose a new meta-algorithm, which we call cGAN-label-recovery, which use any conditional generator $G(z, y)$ trained on the noisy samples, 
to estimate the true label, as $\hat{y}$, of a sample $x$ using the following optimization.
\begin{align} \label{eq:label_recover}
\hat{y} \;\; \in \;\;  \arg \min_{y \in [m]}  \, \big\{ \min_{z_y} \tnorm{G(z_y, y) - x}^2 \; \big\} \,.
\end{align}
%We will compare GANClassifier algorithm \cite{natarajan2013learning} to unbiased classifier in Section \ref{sec:experiments}.
In the right panel of Figure \ref{fig:mnist1} we compare  the {\em label recovery accuracy} of the meta-algorithm using the five conditional GANs, on $500$ randomly chosen noisy training samples.
% If the accuracy is high, the GAN is robust. We compare these accuracies 
This is also compared to a state-of-the-art method \cite{natarajan2013learning} for label recovery, which proposed minimizing unbiased loss function given the noisy labels and the confusion matrix. This unbiased classifier, was shown to outperforms the robust classifiers  \cite{liu2003building, stempfel2007learning, stempfel2009learning} and can be used to predict the true label of the training examples. 
%We see that all RCGAN-classifiers significantly outperforms the unbiased classifier method. 
In Figures \ref{fig:mnist3} of Appendix \ref{app:A}, we show example images from all the generators.

\begin{figure}[ht]
\begin{center}
\includegraphics[width=0.4\textwidth]{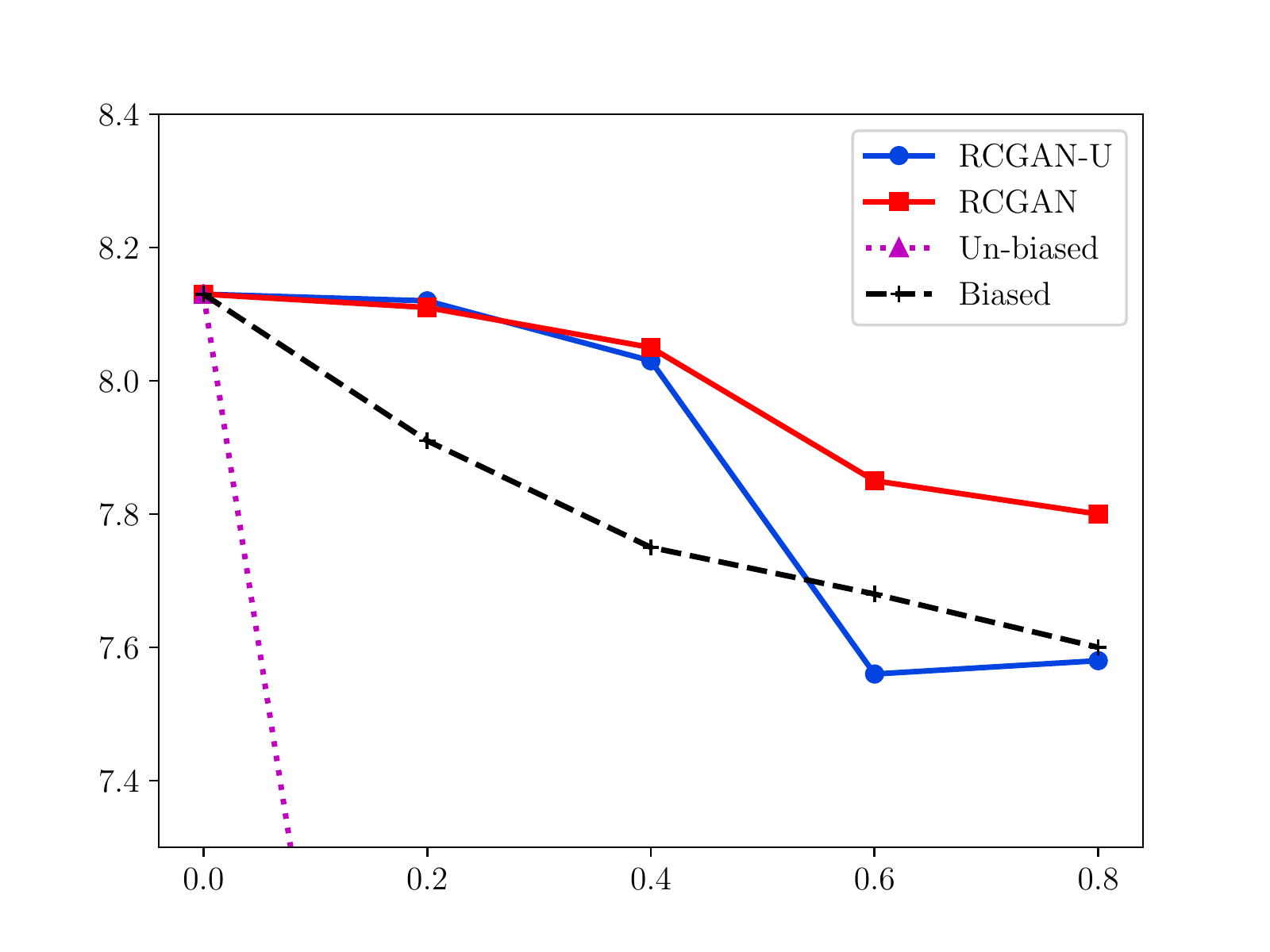}
\put(-151,-5){\small{noise in the real data $(1-\alpha)$}}	
%\put(-160,25){\rotatebox{90}{\small{Inception score}}}	
\put(-180,130){{inception score}}
\includegraphics[width=0.4\textwidth]{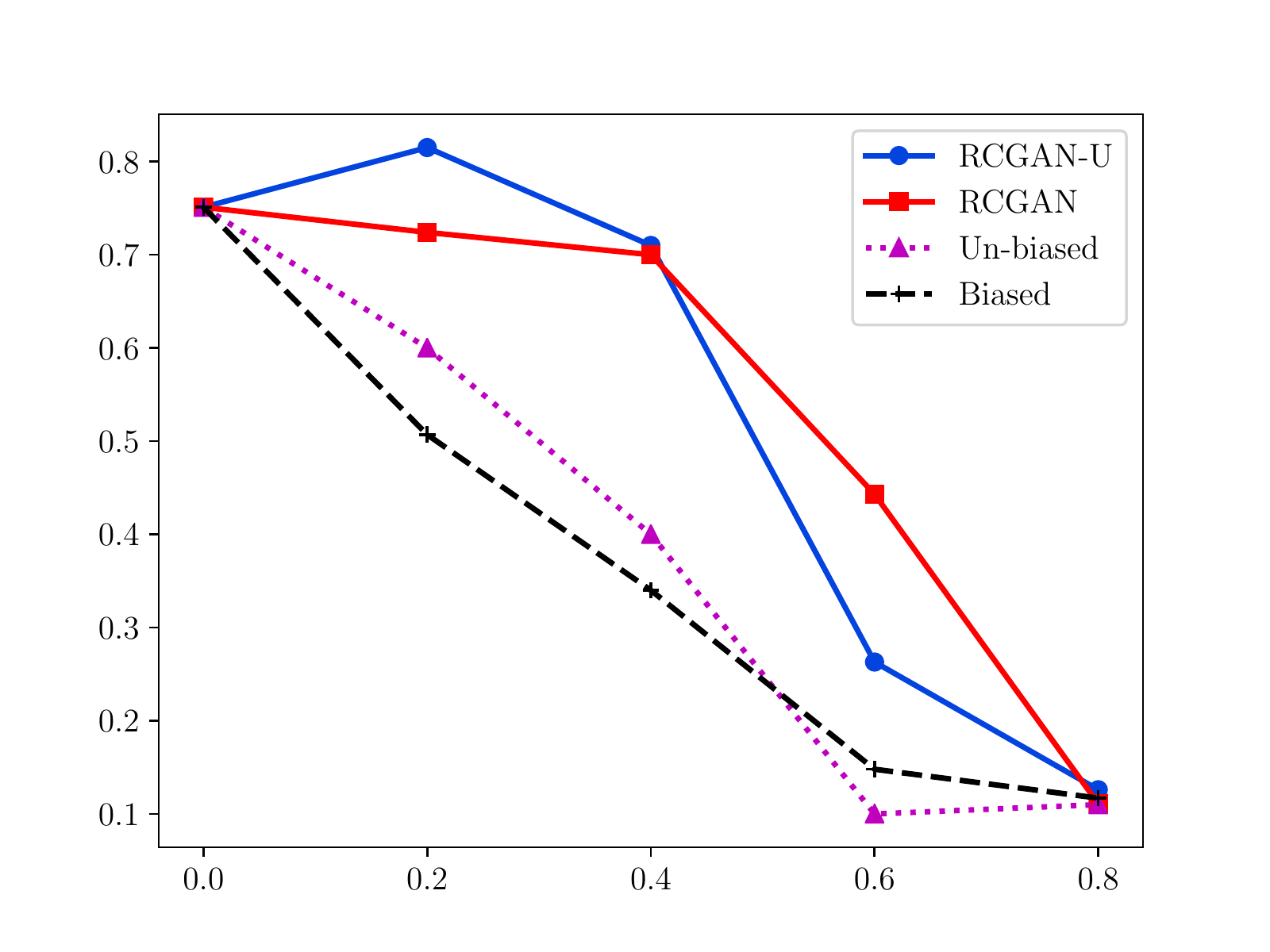}
\put(-151,-5){\small{noise in the real data $(1-\alpha)$}}	
\put(-180,130){{generator label accuracy}}
\caption{Noisy CIFAR-$10$ dataset: Our RCGAN (red) and RCGAN-U (blue) consistently improves upon Unbiased (magenta) and Biased (black) GANs trained on noisy CIFAR-$10$ in inception scores (left) and in generator label accuracy (right). 
The numerical values of the data points are given in a table in Appendix~\ref{app:A}.}
\label{fig:cifar1}
\end{center}
\end{figure}

\subsection{CIFAR-$\mathbf{10}$}
In Figure \ref{fig:cifar1}, we show the inception score  \cite{salimans2016improved} and the label accuracy of the conditional generator for the four approaches: our proposed RCGAN and RCGAN-U, against the baselines Unbiased (Section \ref{sec:exp}) and Biased (Section \ref{sec:intro}) GANs trained using CIFAR-$10$ images \cite{krizhevsky2009learning}, while varying the label accuracy of the real data under uniform flipping. 
In RCGAN-U, even with the regularizer, the learned confusion matrix was a permuted version of the true $C$, possibly because a linear classifier might be too simple to classify CIFAR images. To combat this, we initialized the confusion matrix $M$ to be diagonally dominant (Appendix \ref{app:implementation}).

In the left panel of Figure \ref{fig:cifar1}, our RCGAN and RCGAN-U consistently achieve higher inception scores than the other two approaches. The Unbiased GAN is highly unstable and hence produces garbage images for large noise (Fig. \ref{fig:cifar2}), possibly due to numerical instability of $\normconfusinv$, as noted in \cite{khetan2017learning}. %These results verifies the folklore that conditional GANs produce better quality images.
This confirms that robust GANs not only produce images from the correct class, but also produce better quality images. 
In the right panel of Figure \ref{fig:cifar1}, we report the generator label accuracy (Section \ref{sec:mnist_expt}) on $1$k samples generated by each GAN. We classify the generator images using a ResNet-$110$ model trained to an accuracy of $92.3\%$ on the noiseless CIFAR-$10$ dataset\footnote{\url{https://github.com/wenxinxu/resnet-in-tensorflow}}.
Biased GAN has significantly lower label accuracy whereas the Unbiased GAN has low inception score. In Figure \ref{fig:cifar2} in Appendix \ref{app:A}, we show example images from the three generators for the different flipping probabilities. 
% We believe that the differences in performances would be as stark as the MNIST, if underlying GANs used are better at learning the conditional distribution in a noise-less case. 
We believe that the gain in using the proposed robust GANs will be larger, 
when we train to higher accuracy with larger networks and extensive hyper parameter tuning, with latest innovations in GAN architectures, 
for example \cite{ZGM18,LKFO17,Jol18,KAL17,SBR18}.

%%%%%%%%%%%%%%%%%%%%%%%%%%%%%%%%%%%%%%%%%%%
\section{Numerical comparisons with AmbientGAN \cite{bora2018ambientgan}} 
\label{sec:ambient}

%For the MNIST dataset, all our experiments on RCGAN, RCGAN-U, RCGAN+y, and unbiased GAN use the projection discriminator, whereas the biased GAN uses the AmbientGAN (DCGAN) architecture. We observed that DCGAN (with no projection discriminator) 
%is highly unstable and may not converge if the labels of the real and fake samples are not matched per batch. 
%We can do that for the biased GAN, but for all other GANs we cannot match the labels of the samples because the real true labels are unknown. 
In Table \ref{tab:ambientgan}, we plot the generated label accuracy (as defined in Section \ref{sec:mnist_expt}) 
of RCGAN (which uses the proposed projection discriminator) and AmbientGAN (which uses the 
DCGAN with no projection discriminator) for multiple values of  noise levels ($1-\alpha$). 
%Since AmbientGAN is unstable we report the average accuracy over $9$ runs. 
%Clearly, AmbientGAN (DCGAN) architecture performs poorly for noisy data, 
%hence we only use projection discriminator for our experiments (except for biased GAN).
One of the main reasons for the performance drop of AmbientGAN is that 
without the projection discriminator, training of AmbientGAN is sensitive to 
how the mini-batches are chosen. 
For example, if the distribution of the labels in the mini-batch of the real data is 
different from that of the mini-batch of the generated data, 
then the performance of (conditional) AmbientGAN significantly drops. 
This is critical as we have noisy labels, and matching the labels is in the mini-batch is challenging. 
Our proposed RCGAN provides an architecture and training methods 
for applying AmbientGAN to noisy labeled data, to overcome theses challenges.  
When a projection discriminator is used, as in all our RCGAN and RCGAN-U implementations, 
the performance is not sensitive to how the mini-batches are sampled. 
When a discriminator that is not necessarily a projection discriminator is used, as in our RCGAN+$y$ architecture,
we propose a novel scheduling of the training, which avoids local minima resulting from mis-matched mini-batches (explained in Appendix~\ref{app:implementation}).  
The results are averaged over 10 instances. 

\begin{table}[ht]
\begin{center}
	\begin{tabular}{ l c c c }
	   &\multicolumn{3}{c}{{\bf Noise level (1-$\alpha$)}}\\ \cmidrule(r){2-4}   
	  & 0.2 & 0.3 & 0.5 \\\hline
	    {\bf RCGAN}  &	0.994 & 0.994 & 0.994\\
	   {\bf  AmbientGAN}  & 0.940 & 0.902 & 0.857\\
	\end{tabular}
	\caption{Noisy MNIST dataset: in generated label accuracy, RCGAN improves upon 
	the standard implementation of the AmbientGAN with DCGAN architecture (we refer to Appendix \ref{app:implementation} for implementation details) }
	\label{tab:ambientgan}
\end{center}
\end{table}

%%%%%%%%%%%%%%%%%%%%%%%%%%%%%%%%%%%%%%%%%%%

\section{Conclusion}
\label{sec:conclusion}

Standard conditional GANs can be sensitive to 
noise in the labels of the training data. 
We propose two new architectures to make them robust, 
one requiring the knowledge of the distribution of the noise 
and another which does not,  
and demonstrate the robustness on benchmark datasets of CIFAR-10 and MNIST. 
We further showcase how  the learned generator can be used to 
recover the corrupted labels in the training data, 
which can potentially be used in practical applications.  
The proposed architecture combines the noise adding idea of AmbientGAN \cite{bora2018ambientgan}, 
projection discriminator of \cite{miyato2018cgans}, and 
regularizers similar to those in InfoGAN \cite{CDH16}. 
Inspired by AmbientGAN \cite{bora2018ambientgan}, the main idea is to 
pair the generator output image with  a label that is passed through a noisy channel, 
before feeding to the discriminator. 
We justify this idea of noise adding 
by identifying a certain class of discriminators that have good generalization properties.  
In particular, we prove that projection discriminator, introduced in \cite{miyato2018cgans}, has a good generalization property. 
We showcase that the proposed architecture, 
when trained with a regularizer,  
has superior robustness on benchmark datasets. 
%Our formulation of robust training of GANs from noisy labels poses several interesting research directions. 

One weakness of our theoretical result in Theorem \ref{thm:nn_sample} is that 
depending on the choice of $\cF_{p,L}$ (i.e.~the representation power of the parametric class $D_u(x,y)$),  
closeness in the neural network distance 
does not always imply closeness of the distributions. 
It is generally a challenging problem to address generalization for specific 
function class $\cF$ and a pair of distributions $P$ and $Q$. 
However, a recent breakthrough in generalization properties of GAN in \cite{BMR18} 
makes the connection between 
 $d_{\cF}(\tP_n,\tQ_n)$ and $d_{\rm TV}(P,Q)$ precise, under some assumptions on the $P$ and $Q$.  
This leads to the following research question: 
under which class of distributions $P$ and $Q$ does 
the neural network distance of the 
proposed conditional GAN with projection discriminator generalize? 
The emphasis is in studying the class of functions satisfying Assumption \ref{eq:mix_inv} and 
identifying corresponding family of distributions that generalize under this function class. 

\section*{Acknowledgement}
This work is supported by NSF awards CNS-1527754, CCF-1553452, CCF-1705007, RI-1815535 and Google Faculty Research Award.
This work used the Extreme Science and Engineering Discovery Environment (XSEDE), which is supported by National Science Foundation grant number OCI-1053575.  Specifically, it used the Bridges system, which is supported by NSF award number ACI-1445606, at the Pittsburgh Supercomputing Center (PSC).
%We are grateful to Amazon for providing us 
This work is partially supported by 
the generous research credits on AWS cloud computing resources from Amazon. 

%%%%%%%%%%%%%%%%%%%%%%%%%%%%%%%%%%%%%%%%%%%%%%%%%%%%%%%%

% -------------------------------------------------------------------------------------------------------------------------------------------------------------------------------------------------

\bibliography{gan,_gan}

% -------------------------------------------------------------------------------------------------------------------------------------------------------------------------------------------------
\clearpage
\appendix
\section*{Appendix}

\section{Notations and Lemmas}

\subsection{Additional Notation}
Here we define some additional notations required for the proof. We define certain notations before we provide the main theoretical contributions of our paper. If $f(x,y)$ is a function of two variable of $x, y$, where $y \in [\ny]$, then $\vect{f}(x)$ is the vector $[f(x, 1), \cdots, f(x, \ny)]^T$. If $\xy{P}$ is probability distribution of $(X, Y) \in \cX \times [\ny]$, then $\ygx{P}$ is the conditional distribution of $Y$ given $X=x$.

For a matrix $A$, let $\lnorm{A}{p} = \max_{\lnorm{x}{p} = 1} \lnorm{Ax}{p}, \forall p \in \naturals \cup \{0, \infty\}$. Then $\maxnorm{A} = \max_i \sum_{j} \abs{A_{ij}}$, $\onorm{A} = \max_j \sum_{i} \abs{A_{ij}}$  and  $\tnorm{A} = \sigma_{\max}(A)$, the maximum singular value. $\ones$ is all ones vector with appropriate dimensions and $\identy$ is identity matrix with appropriate dimensions. $[n] = \{1,2,\ldots,n\}\;, \forall n \geq 1$. For a vector $x \in \reals^n$, $x_i$ ($i \in [n]$) is its $i$-th coordinate.

For the sake of proof we will assume that $\cF$ is class of vector functions of the form $\D(x) \in \reals^\ny$. In terms of the notation in the main material original $\D(x, y)$ is $\D(x)_y$ here.
For a class $\cF$ of vector valued functions $\D: \cX \to \reals^{n}$. Therefore we re-define the operation $\vopsymb$ between a matrix $T \in \reals^{n \times n}$ and $\cF$ as,
\begin{align*}
%\label{eq:vop}
\vop{T}{\cF} = \{T\D(\cdot) \,|\, f(\cdot) \in \cF \}\,. 
\end{align*}

If $\xy{P}$ is probability distribution of $(X, Y) \in \cX \times [\ny]$, then $\ygx{P}$ is the conditional discrete distribution of $Y$ given $X=x$, $\x{p}(x)$ is the marginal density of $X$, and  
\begin{align}
\vygxP &= [\ygx{P}(Y=1),\, \ygx{P}(Y=1),\, \ldots,\, \ygx{P}(Y=\ny)]^T \text{, and} \label{eq:vec_pmf} 
\\
\vect{\x{p}}(x) &= \x{p}(x)\vygxP \label{eq:vec_density}
\end{align}

% -------------------------------------------------------------------------------------------------------------------------------------------------------------------------
\subsection{Supporting Lemmas}
\begin{lemma}[Characterization of \NND] \label{lem:metric}
$\nnd{P}{Q} \geq 0$ for all $P, Q$. And if $\phi$ is a convex or concave function, then the Neural network distance is $0$ when the distributions are same, i.e. $\nnd{P}{P} = 0$.
\begin{proof}
For concave $\phi(\cdot)$ we define $\mu_\phi = \phi(1/2)$. Since, by definition $\D=1/2\, \ones$ is feasible solution to the optimization problem in \eqref{eq:nnd}, thus $\nnd{P}{Q} \geq 0$.
\begin{align}
\nnd{P}{P} = &\sup_{\D \in \cF} \;\; \expectD{(x,y) \sim P}{\phi\left(\D(x)_y\right)} + \expectD{(x,y) \sim P}{\phi\left(1 - \D(x)_y\right)} - 2\phi(1/2) \nonumber \\
\leq &\sup_{\D \in \cF} \;\; 2\;\phi\left( \expectD{(x,y) \sim P}{\frac12(\D(x)_y + 1 - \D(x)_y)} \right) - 2\phi(1/2) \nonumber \\
= &\sup_{\D \in \cF} \;\; 2\;\phi\left( 1/2 \right) - 2\phi(1/2)  = 0 \nonumber 
\end{align}
The inequality in second line follows from Jensen's inequality for concave $\phi(\cdot)$.

For convex $\phi(\cdot)$ we define $\mu_\phi = \phi(0) + \phi(1)$. Since, by definition $\D=1$ is feasible solution to the optimization problem in \eqref{eq:nnd}, thus $\nnd{P}{Q} \geq 0$.
\begin{align}
\nnd{P}{P} &= \sup_{\D \in \cF} \;\; \expectD{(x,y) \sim P}{\phi\left(\D(x)_y\right)} + \expectD{(x,y) \sim P}{\phi\left(1 - \D(x)_y\right)} -\phi(0) -\phi(1) \nonumber \\
&= \sup_{\D \in \cF} \;\; \expectD{(x,y) \sim P}{\phi\left(\D(x)_y\right) + \phi\left(1 - \D(x)_y\right)} - (\phi(0) + \phi(1))\nonumber \\
&\leq \sup_{\D \in \cF} \;\; \expectD{(x,y) \sim P}{\phi(0) + \phi(1)} - (\phi(0) + \phi(1)) = 0\nonumber
\end{align}
The last inequality follows from Jensen's inequality for convex $\phi(\cdot)$
\end{proof}
\end{lemma}
This Lemma \ref{lem:metric} ensures that all the multiplicative lower bounds and upper bounds in Theorem \ref{thm:gen_counter} and its corollaries implies recoverability.

\begin{lemma} \label{lem:noise_prob}
If $P$ is a distributions on $\cX \times [\ny]$ and $\tP$ is the distribution of sample $(X, \tY)$ of $P$ when passed through the noisy-channel given by the confusion matrix $\confus \in \reals^{\ny \times \ny}$ (as defined in Section \ref{sec:noise-model}). Then,
\begin{align}
\vtygxP = \confus^T \vygxP\,,
\end{align}
where $\vygx{P} = [\ygx{P}(Y=1),\, \ygx{P}(Y=1),\, \ldots,\, \ygx{P}(Y=\ny)]^T$.
\begin{proof}
\begin{align*}
\tygx{\tP}(\tY=j) &= \sum_{i \in [\ny]}\prob{\tY=j | Y=i} \ygx{P}(Y=j),\, \forall j \in [\ny] \\
\tygx{\tP}(\tY=j) &= \sum_{i \in [\ny]} \confus_{ij} \ygx{P}(Y=j),\, \forall j \in [\ny] \\
\vtygxP &= \confus^T \vygxP
\end{align*}
\end{proof}
\end{lemma}

% -------------------------------------------------------------------------------------------------------------------------------------------------------------------------
\section{Proof of Theorem \ref{thm:prob_dist_up_lb}} 
	\label{app:prob_dist_up_lb_pf} 
	We first prove the approximation bounds for total variation distance in Eq.~\eqref{eq:tv_ub_lb}, and then use it to prove similar bounds for the Jensen-Shannon divergence in Eq.~\eqref{eq:js_ub_lb}. 
	Recall that total variation distance between $P$ and $Q$ can be written in several ways: 
	\begin{eqnarray*}
		d_{\rm TV} (P,Q) &=& \max_{S_1,\ldots,S_m} \sum_{y\in[m]}  P(S_y,y)-Q(S_y,y) \\
			&=& \max_{S_1,\ldots,S_m} \sum_{y\in[m]}  | P(S_y,y)-Q(S_y,y) | \\
			&=& \max_{S_1,\ldots,S_m} \| P(\{S_y\}_{y\in[m]},\cdot) - Q(\{S_y\}_{y\in[m]},\cdot) \|_1 \;,
	\end{eqnarray*}
	where we used the notation of a row-vector 
	$P(\{S_y\}_{y\in[m]},\cdot) = [ P(S_1,1), \cdots, P(S_m,m) ] $. The lower bound on $d_{\rm TV}(\tP,\tQ)$ follows that 
\begin{eqnarray*}
	d_{\rm TV} (P,Q) &=& \max_{S_1,\ldots,S_m \subseteq \cX} \sum_{y\in[m]}  \big\{ P(S_y,y) - Q(S_y,y) \big\} \\
		&=& \max_{S_1,\ldots,S_m \subseteq \cX}  \langle  \,  \ones , P(\{S_y\}_{y\in[m]},\cdot)- Q(\{S_y\}_{y\in[m]},\cdot)  \, \rangle \\
		&\overset{(a)}{=}& \max_{S_1,\ldots,S_m \subseteq \cX}  \langle  \,  \ones ,  \big( \tP(\{S_y\}_{y\in[m]},\cdot) - \tQ(\{S_y\}_{y\in[m]},\cdot) \big)C^{-1}  \, \rangle \\
		&\overset{(b)}\leq& \vertiii{C^{-T}}_1 \;  \max_{S_1,\ldots,S_m \subseteq \cX}   \big\| \tP(\{S_y\}_{y\in[m]},\cdot) - \tQ(\{S_y\}_{y\in[m]},\cdot) \big\|_1 \\
		&\overset{(c)}{=}&  \vertiii{C^{-1}}_\infty \, d_{\rm TV} (\tP,\tQ) \;,
\end{eqnarray*}
where $(a)$ follows from the fact that $\tP(\{S_y\}_{y\in[m]}, \cdot ) = P(\{S_y\}_{y\in[m]}, \cdot )\,C$, 
$(b)$ follows from the fact that $\ones^T A x \leq \|Ax\|_1 \leq \vertiii{A}_1\|x\|_1$, and 
$(c)$ follows  from $\vertiii{A}_1=\vertiii{A^T}_\infty$. 
%\begin{align}
%\tv{P}{Q}
%%&= \frac12 \int_{\cX \times [\ny]} \abs{dP-dQ} \nonumber \\
%&= \frac12 \int_{\cX} \sum_{y \in [\ny]} \abs{p(x)\ygx{P}(y)-q(x)\ygx{Q}(y)} dx \nonumber \\
%&\overset{(a)}{=}  \frac12 \int_{\cX} \onorm{p(x)\vygxP-q(x)\vygxQ} dx \nonumber \\
%&\overset{(b)}{=} \frac12 \int_{\cX} \onorm{\tp(x) \confus^{-T}\vtygxP - \tq(x)  \confus^{-T} \vtygxQ} dx \nonumber \\
%&\overset{(c)}{\leq} \onorm{\confus^{-T}} \frac12 \int_\cX \onorm{\tp(x) \vtygxP- \tq(x) \vtygxQ} dx \nonumber \\
%&= \maxnorm{\confus^{-1}} \tv{\tP}{\tQ} \nonumber
%\end{align}
%where $(a)$ uses the definition in equation \eqref{eq:vec_density}, $(b)$ follows from Lemma \ref{lem:noise_prob}, and $(c)$ follows from the definition of induced matrix L-$1$ norm (Section \ref{sec:notations}).
%Finally to get the desired result we use the fact that $\onorm{C^{-T}} = \maxnorm{C^{-1}}$ to get the lower bound. 
The upper bound follows from similar arguments: 
\begin{align}
\tv{\tP}{\tQ} %&= \frac12 \int_{\cX} \onorm{\tp(x) \vtygxP- \tq(x) \vtygxQ} dx \nonumber \\
%&\overset{(b)}{=} \frac12 \int_{\cX} \onorm{p(x) \confus^{T} \vygxP- q(x) \confus^{T} \vygxQ} dx \nonumber \\
&\;\; \leq \;\; \vertiii{\confus^{T}}_1 \max_{S_1,\ldots,S_m \subseteq \cX}   \big\| P(\{S_y\}_{y\in[m]},\cdot) - Q(\{S_y\}_{y\in[m]},\cdot) \big\|_1   \nonumber \\
&\;\;  =\;\;    \tv{P}{Q} \nonumber
\end{align}
where last equality uses the fact that $\onorm{C^T} = 1$ for all row-stochastic matrices $C$. 

To prove the approximation bounds for Jensen-Shannon divergence, 
we use the following lemma that bounds the JS divergence by the TV distance. 
\begin{lemma} 
\label{lem:js_tv}
$\frac12 \tv{P}{Q}^2 \;\; \leq \;\; \js{P}{Q} \;\; \leq \;\; 2 \,\tv{P}{Q}$. 
\end{lemma} 
A proof is provided in Section \ref{sec:proof_pinsker}. 
The following series of inequalities follow from this lemma. 
%\noindent 
%\vspace{-2.cm}
\begin{alignat*}{2}
\frac12\, \tv{\tP}{\tQ}^2  \;\; &\overset{(a)}\leq \;\; \js{\tP}{\tQ} \;\; && \overset{(a)}\leq \;\; 2\, \tv{\tP}{\tQ} \\
\frac{\normconfust}2\, \tv{P}{Q}^2  \;\; &\overset{(b)}\leq  \;\; \js{\tP}{\tQ}  \;\; &&\overset{(b)}\leq \;\; 2\, \tv{P}{Q} \\
\frac{\normconfust}{8}\, \js{P}{Q}^2  \;\; &\overset{(a)}\leq    \;\; \js{\tP}{\tQ} \;\; &&\overset{(a)}\leq  \;\; \sqrt{8\,\js{P}{Q}} 
\end{alignat*}
where $(a)$ uses Lemma \ref{lem:js_tv}, and $(b)$ uses equation \eqref{eq:tv_ub_lb}.

% -------------------------------------------------------------------------------------------------------------------------------------------------------------------------
\subsection{Proof of Lemma~\ref{lem:js_tv}}
\label{sec:proof_pinsker}

\begin{align}
\kl{P}{\frac{P+Q}2} &= \expectD{X \sim P}{\log \frac{2 P(X)}{{ P(X) + Q(X)}}} \nonumber \\
&\overset{(a)}\leq \expectD{X \sim P}{\frac{2P(X) -({P(X)+Q(X)})}{{ P(X) + Q(X)}}} \nonumber \\
&\leq \sup_X \frac{2 P(X)}{ P(X) + Q(X)} \cdot \expectD{X \sim P}{\frac{\abs{2 P(X) -({P(X)+Q(X)})}}{{2 P(X)}}} \nonumber \\
&\leq 2 \cdot 2\tv{P}{\frac{P+Q}2} \nonumber\\
&\leq 2\, \tv{P}{Q} \label{eq:kl_ub_tv}\,,
\end{align}
where $(a)$ uses $\log x \leq x-1$.
\noindent
%\vspace{-.5cm}
\begin{align*}
\js{P}{Q} &= \frac12 \kl{P}{\frac{P+Q}2} + \frac12 \kl{Q}{\frac{P+Q}2} \\ %\overset{(a)}\leq 2 \tv{P}{Q} \\
&\overset{(b)}\geq \tv{P}{\frac{P+Q}2}^2 + \tv{Q}{\frac{P+Q}2}^2 \\
&= \frac12 \tv{P}{Q}^2
\end{align*}
where $(a)$ uses equation \eqref{eq:kl_ub_tv}, and $(b)$ uses Pinsker's inequality $0.5\,\kl{P}{Q} \leq \tv{P}{Q}^2$

%The lower bound follows from the Pinsker's inequality, $2 d_{\rm TV}(P,Q)^2 \leq d_{\rm KL}(P\|Q) $, 
%and the definition of JS divergence, $d_{\rm JS}(P\|Q) = (1/2)d_{\rm KL}(P|(P+Q)/2) + (1/2)d_{\rm KL}(Q|(P+Q)/2)$. 
%Together we get that 
%$d_{\rm JS}(P\|Q) \geq 2 d_{\rm TV}(P,(P+Q)/2) \geq (1/2)d_{\rm TV}(P,Q) $.

% -------------------------------------------------------------------------------------------------------------------------------------------------------------------------------------------------
\section{Examples satisfying Assumption~\ref{eq:mix_inv}}
\label{sec:example}

Several classes of functions that are used in practice and studied in theory indeed satisfy 
our Assumption~\ref{eq:mix_inv}. 
For example, consider the set of 1-Lipschitz continuous and bounded functions 
$\cF=\{ f:\reals^d \times [m]\to \reals \,|\, 0\leq f(x,y) \leq 1 \text{ for all $x$ and $y$, and } | f(x_1,y_1) - f(x_2,y_2)| \leq \|x_1 - x_2\| + |y_1-y_2|\text{ for all $x_1,x_2$, $y_1$, and $y_2$}   \}$, 
which was studied in \cite{arora2017generalization} in the context of generalization bounds for the neural network distance.
It follows that this $\cF$ satisfies Assumption \ref{eq:mix_inv}, with $c=1/2$ 
and $ \cF- 1/2 \in \cF_1$.  
This is a special case of some examples of classes of functions satisfying the assumption, 
that we provide in the following Remark.
A proof is provided in Appendix \ref{app:proj_example_pf}.

\begin{remark} 
	\label{rem:proj_example}
	The following classes of discriminators satisfy the inclusion condition in Assumption \ref{eq:mix_inv}: 
	\begin{itemize}
		\item[1.] Class of all bounded functions $D:\cX\times[m]\to[c_1, c_2]$ for any $c_1\leq c_2 \in \reals$.
		\item[2.] Class of all bounded functions $D:\cX\times[m]\to[c_1, c_2]$, which are $L$-Lipschitz in $x$ for any $c_1\leq c_2 \in \reals$ and $L\geq0$.
		\item[3.] Class of all bounded functions $D:\cX\times[m]\to[c_1, c_2]$, which are $L$-Lipschitz in $x$ and $y$ for any $c_1\leq c_2 \in \reals$ and $L\geq(c_2-c_1)$.
%		\item[4.] $\cF = \{ D:\cX\times[m] \to \reals \,|\, D(x,y)= {\rm vec}(y)^T \, \Theta_0\, \psi(x; \Theta_1), \Theta_0 \in \reals^{m\times m}\}$, for any parametric family $\psi(x; \Theta_1)$ parametrized by $\Theta_1$, and one-hot encoding of ${\rm vec}(y)^T = [{\mathbb I}_{y=1} , \ldots, {\mathbb I}_{y=m} ]$.
	\end{itemize}
\end{remark}

% -------------------------------------------------------------------------------------------------------------------------------------------------------------------------------------------------
\section{Proof of Remark \ref{rem:up_lb_tight}: the tightness of Theorem \ref{thm:gen_ub_lb}}
\label{sec:tight}

%\section{Proof of Remark \ref{rem:up_lb_tight}}
%\label{app:up_lb_tight_pf}
Let $\cX = \{x_1, x_2\}$ and $\cF =\{f \,|\, f(x, y) \in [-1, 1]\}$. From Remark \ref{rem:proj_example}, we know that $\cF$ satisfies Assumption \ref{eq:mix_inv}. Then it is easy to check that $\nndff{\cF}{P}{Q} = \onorm{(P-Q)((\{x_1\}, \cdot)} + \onorm{(P-Q)((\{x_2\}, \cdot)}$, where we used the notation $P(\{x\},\cdot) = [ P(\{x\},1), \cdots, P(\{x\},m) ] $.
\begin{enumerate}
	\item {\bf Lower bound:} It is easy to show that there exists $P_{X, Y}$ and $Q_{X, Y}$ such that $(P-Q)((\{x_1\}, \cdot) = \epsilon \ones$ and $(P-Q)((\{x_2\}, \cdot) = -\epsilon \ones$ for any $\epsilon \in [0, 1/m]$. Thus,
	\begin{align}
		\nndff{\cF}{P}{Q} = \onorm{ \epsilon \ones} + \onorm{- \epsilon \ones} = 2 m \epsilon \,.
	\end{align}
	Then we can show that,
	\begin{align}
		(\tP-\tQ)(\{x\}, \cdot) = C^T (P-Q)(\{x\}, \cdot) = \pm \epsilon\, C^T \ones\,
	\end{align}
	Thus,
	\begin{align}
		\nndff{\cF}{\tP}{\tQ} = 2\onorm{\pm \epsilon C^T\ones} = 2 \epsilon \, \ones^T C^T\ones = 2 \epsilon \, \ones^T \ones  = 2 m \epsilon  \,,
	\end{align}
	where the penultimate equality follow from $C \ones = \ones$, since $C$ is row-stochastic.
	
	\item {\bf Upper bound:} We can again show that there exists $P_{X, Y}$ and $Q_{X, Y}$ such that $(P-Q)((\{x_1\}, \cdot) = \epsilon C^{-T} v$ and $(P-Q)((\{x_2\}, \cdot) = -\epsilon C^{-T} v$, where $v \in {\arg\max}_{v_i \in \pm 1} \onorm{C^{-T} v}$, for sufficiently small $\epsilon$. Thus,
	\begin{align}
		\nndff{\cF}{P}{Q} = 2\onorm{\pm \epsilon C^{-T} v} = 2 \epsilon \onorm{C^{-T}}\onorm{v} = 2 \epsilon \maxnorm{C^{-1}}\onorm{v}\,.
	\end{align}
	Using similar steps as the above lower-bound case, we can show that,
	\begin{align}
		\nndff{\cF}{\tP}{\tQ} = 2\onorm{\pm \epsilon C^TC^{-T} v} = 2 \epsilon \onorm{v}\,.
	\end{align}
\end{enumerate}

% -------------------------------------------------------------------------------------------------------------------------------------------------------------------------
\section{Proof of Theorem \ref{thm:gen_ub_lb}} 
\label{app:gen_ub_lb_pf}

We start with the following two key lemmas. 
\begin{lemma} For any sample $(X,Y) \sim P$ (or $Q$), 
	let $\tP$ (or $\tQ$) denote the sample whose label $Y$ is corrupted by 
	a noise defined by a confusion matrix $C$, where $C_{y \ty}$ is the probability the 
	a label $y$ is corrupted as a label $\ty$. Then, for any $P$, $Q$, and any class of discriminators $\cF$, 
	\label{lem:transform_nnd}
	\begin{align}
		\nndff{\cF}{\tP}{\tQ} &\;\; = \;\; \nndff{\vop{C}{\cF}}{P}{Q}  \;.
		\label{eq:transform_nnd}
	\end{align}
\end{lemma}

\begin{lemma} If a class of discriminators $\cF$ satisfies Assumption \ref{eq:mix_inv} with a constant shift $c$, then 
for all row-stochastic non-negative matrix  $C$, 
	\label{lem:subset}
	\begin{eqnarray}
		\vop{\normconfus \identy}{(\cF-c)} \;\; \subseteq \;\; \vop{\confus}{(\cF-c)} \;\; \subseteq \;\; (\cF-c)\;.
		\label{eq:subset}
	\end{eqnarray}
\end{lemma}

From the ordering of sets of functions in  Lemma \ref{lem:subset}, it follows that 
\begin{align}
	\nndff{\vop{\confus}{\cF}}{P}{Q} &\;\; \geq \;\; \nndff{\vop{\normconfus \identy}{\cF}}{P}{Q}    \;\; = \;\; \normconfus \nndff{\cF}{P}{Q}  \label{eq:gen_up_lo_bd}
\end{align}
where the last equality follows from the fact that $\D \in \cF$ if and only if $\alpha \D \in \vop{\alpha \identy}{\cF}$. 
Note that we ignored the shift $c$, as  the definition of the neural network distance is invariant to any 
constant shift $c$, and we can cancel any given shift $c$ of a set $\cF$ as we see fit. 
Next, it follows immediately from Lemma \ref{lem:subset} that 
\begin{align}
	 \nndff{\vop{\confus}{\cF}}{P}{Q} \;\; \leq \;\; \nndff{\cF}{P}{Q} \;.
\end{align}
This finishes the proof of the theorem. 

% -------------------------------------------------------------------------------------------------------------------------------------------------------------------------
\subsection{Proof of Lemma~\ref{lem:transform_nnd}} 
\label{sec:proof_transform_nnd}

For any function $D: \cX \times [m] \to \reals$ and a distribution $P$ over $\cX\times [m]$, 
we denote the expectation by  $\langle D , P \rangle = \E_{(X,Y)\sim P}[D(X,Y)]$. 
Further, we let 
$\tP(S,\ty) = \sum_y P(S,y) C_{y \ty} = PC(S,\ty)$ denote the distribution of the corrupted sample, by noise with confusion matrix $C$.  
Note that we intentionally overloaded the matrix multiplication notation $PC$. 
We also treat $D$ as a (infinite-dimensional) matrix, and let $DC^T(x,y)= \sum_{\ty} D(x,\ty)C_{y\ty}$. 
It follows that, 
\begin{eqnarray*}
	d_\cF(\tP,\tQ) &\triangleq& \sup_{D\in \cF} \;\; \E_{\tP}[D(X,Y)] - \E_{\tQ}[D(X,Y)] \\
			     &=& \sup_{D\in \cF} \la D, \tP-\tQ\ra \\
			     &=& \sup_{D\in \cF} \la D, (P-Q)C \ra \\
			     &=& \sup_{D\in \cF} \la D C^T, (P-Q) \ra \\
			     &=& \sup_{\tD\in \vop{C}{\cF}} \la \tD , (P-Q) \ra \;,\\
\end{eqnarray*}
where the last equality follows from the definition of the $\circ$ operation in Eq.~\eqref{eq:vop}. 
This proves the desired claim.

\subsection{Proof of Lemma~\ref{lem:subset}}

		We will prove the desired claim for $\cF_1$ and $\cF_2$ separately. 
		As the set orderings are preserved under addition of sets, 
		this proves the desired claim.  
		
		First we will prove the Lemma when condition $\vop{T}{\cF} \subseteq \cF$ holds. 
		We use the notations from Appendix\ref{sec:proof_transform_nnd}. 
		We want to prove,
		\begin{align*}
		&\vop{\normconfus \identy}{\cF} \subseteq \vop{\confus}{\cF} \\
		\iff &\forall \D \in \cF,\; \exists \D' \in \cF \text{ such that } \normconfus \D(x,y) =  \D'C^T(x,y),\; \;\;\; \forall x \in \cX, \forall y\in[m] \\
		\iff &\forall \D \in \cF,\; \exists \D' \in \cF \text{ such that } \normconfus  \D C^{-T}(x,y) = \D'(x,y),\;\;\;\; \forall x \in \cX, \forall y\in[m] \\
		\iff &\vop{\normconfus \confus^{-1}}{\cF} \subseteq \cF\;. 
		\end{align*}
		The last statement is true by the the Assumption \ref{eq:mix_inv} because $\maxnorm{\left(\confus^{-1} / \maxnorm{\confus^{-1}}\right)} = 1$. The second covering $\vop{\confus}{\cF} \subseteq \cF$ is also true by Assumption \ref{eq:mix_inv} because $\maxnorm{\confus} = 1$ since $\confus$ is row-stochastic matrix with rows summing to $1$.
		
		Now we will prove the case when  condition $\cF = \{\alpha f(x) \,|\, f(x)_y = g(x) \in \reals, g \in \cF_x  , \alpha\in[0,1]\}$ holds. $\vop{C}{\cF} \subseteq \cF$ holds true because $\ones$ is an eigenvector of $\confus$ with eigenvalue $1$. Similarly to the previous condition, we need to prove that $\vop{\confus^{-1}/\normconfusinv}{\cF} \subseteq \cF$, but since $\ones$ is an eigenvector of $C^{-1}$ with eigenvalue $1/\normconfusinv \leq 1$, the again holds true.

% -------------------------------------------------------------------------------------------------------------------------------------------------------------------------
% -------------------------------------------------------------------------------------------------------------------------------------------------------------------------

\section{Proof of Remarks \ref{rem:proj_example} and \ref{rem:proj}}
\label{app:proj_example_pf}

% -------------------------------------------------------------------------------------------------------------------------------------------------------------------------
\subsection{Class of all bounded functions (Total Variation)}
Let $\cF([c_1, c_2])$ be class of all functions with range inside $[c_1, c_2]^\ny$. Proof follows from the Appendix \ref{app:lip_ub_lb_pf} by taking $\lim_{L \to \infty}$.

% -------------------------------------------------------------------------------------------------------------------------------------------------------------------------
\subsection{Class of all bounded and Lipschitz functions in $x$} \label{app:lip_ub_lb_pf}
Let $\cF(L, [c_1, c_2])$ be class of all vector valued $L$-Lipschitz functions in $x$ with range inside $[c_1, c_2]^\ny$. That is,
\begin{align}
\maxnorm{D(x_1) - D(x_2)} \leq L\tnorm{x_1 - x_2} \text{   and    } D(x) \in [c_1, c_2]\;\; \forall x, x_1, x_2 \in \cX \,, y \in [\ny]
\end{align}
\begin{lemma}
	\label{lem:lip_scale_equiv}
	$\nndff{\cF(L, [c_1, c_2])}{P}{Q} = \frac{(c_2 - c_1)}{2}\;\nndff{\cF(2L/(c_2 - c_1), [-1, 1])}{P}{Q}$
	\begin{proof} There exists a bijection $f: \cF(L, [c_1, c_2]) \to \cF(2L/(c_2-c_1), [-1, 1])$, such
		that $f(D) = \frac{2D-(c_1+c_2)}{c_2 - c_1}$. This is true since $f$ is invertible, $f(D) \in [-1, 1]$, and $f$ is $2L/(c_2 - c_1)$-Lipschitz.
		
		\begin{align}
		\nndff{\cF(L, [c_1, c_2])}{P}{Q} &= \sup_{D \in \cF(L, [c_1, c_2])} \;\; \expectD{(x,y) \sim P}{D(x,y)} - \expectD{(x,y) \sim Q}{D(x,y)} \nonumber \\
		&= \frac{(c_2 - c_1)}{2} \sup_{D \in \cF(2L/(c_2 - c_1), [-1, 1])} \;\; \expectD{(x,y) \sim P}{f(D)(x,y)} - \expectD{(x,y) \sim Q}{f(D)(x,y)} \nonumber \\
		&= \frac{(c_2 - c_1)}{2} \nndff{\cF(2L/(c_2 - c_1), [-1, 1])}{P}{Q}
		\end{align}
	\end{proof}
\end{lemma}

By Lemma \ref{lem:lip_scale_equiv} any $[c_1, c_2]$ is similar to $[-1, 1]$ up to a scaling, thus we only prove for $[-1, 1]$. Now we will show that the inclusion condition (Assumption \ref{eq:mix_inv}) hold for this class of functions.
\begin{lemma}
	\label{lem:lip_subset}
	$\vop{T}{\cF} (L, [-1, 1]) \subseteq \cF (L, [-1, 1])$, $\forall \maxnorm{T} = 1$ 
	\begin{proof}
		We want to show that, $\forall \; D' \in  \vop{T}{\cF} (L, [-1, 1])$, we also have $D' \in \cF (L, [-1, 1])$. In other words, $\forall \; D \in  \cF (L, [-1, 1]),  T {D} \in \cF (L, [-1, 1])$. First the we show that range of $D \in \vop{T}{\cF} (L, [-1, 1])$ is $[-1, 1]$, ie.,
		\begin{align}
		&\maxnorm{T {D}(x)}
		\leq \maxnorm{T} \maxnorm{{D}(x)} \leq 1 \cdot 1  \leq 1\,.
		\end{align}
		In a similar way we prove the Lipschitz property.
		\begin{align}
		\maxnorm{T{D}(x_1) - T{D}(x_2)} &\leq \maxnorm{T} \maxnorm{{D}(x_1) - {D}(x_2)} \leq 1 \cdot L \tnorm{x_1 - x_2}
		\end{align}
	\end{proof}
\end{lemma}

Finally, we can use Theorem \ref{thm:gen_ub_lb} to get the desired result.

% -------------------------------------------------------------------------------------------------------------------------------------------------------------------------
\subsection{Class of all bounded and Lipschitz functions in $x$ and $y$}

As $|y_1-y_2| \geq 1$ for all $y_1 \neq y_2$, 
$L$-Lipschitz functions with $L\geq c_2-c_1$ only imposes 
conditions on pairs of data with the same values of $y$.  
Hence, this boils down to the previous case studied in Appendix~\ref{app:lip_ub_lb_pf}.

% -------------------------------------------------------------------------------------------------------------------------------------------------------------------------
\subsection{Class of projection discriminators}
%\subsection{Class of Projection Discriminators}
\label{sec:proof_assumption_proj}
The function class is $\cF = \{f_1 + f_2 \,|\, f_1 \in \cF_1^{(\theta)}, f_2 \in \cF_2^{(\theta)}, \theta \in \reals^{d_\theta} \}$, where
\begin{align}
&\cF_1^{(\theta)} = \big\{ \, {\rm vec}(y)^T \, V\,  \psi(x;\theta) \,\big|\, V\in\cV_1\, \big\}\text{, and,} \\
&\cF_2^{(\theta)} = \big\{ \, v^T\,\psi'(x;\theta) \,\big|\, v \in\cV_2  \, \big\}\,\text{, where,}\\
&\cV_1 = \big\{ \, V\in\reals^{m\times d_V} \,\big|\, \max_{i} |V_{ij}| \leq 1 \text{ for all } j\in[d_V]  \, \big\}  \;,
%&\cV_1 = \big\{ \, V\in\reals^{m\times d_V} \,\big|\, \max_{i} \sum_{j} |V_{ij}| \leq 1 \, \big\}  \;,
\label{eq:cond1_proj}
\text{ and }\\
&\cV_2 = \big\{ \, v \in \reals^{d_v} \,\big|\, \|v\| \leq 1 \, \big\}\;.
\end{align}
We will show that both $\cF_1^{(\theta)}$ and $\cF_2^{(\theta)}$ satisfy Assumption \ref{eq:mix_inv}. For any $A \in \reals^{m\times m}$, we can write $A \circ \cF_1^{(\theta)}$ as 
\begin{align}
A \circ \cF_1^{(\theta)} = \{ {\rm vec}(y)^T A V \psi(x;\theta) \, \big|\, V \in \cV_1 \, \}\;.
\end{align}
If $\maxnorm{A} = 1$, then, $\max_{i \in [m]} |(AV)_{ij}| = \maxnorm{(AV)_{\cdot,j}} = \maxnorm{A(V_{\cdot,j})} \leq \maxnorm{A} \maxnorm{V_{\cdot,j}} = 1 \cdot \max_{i \in [m]} |V_{ij}| \leq 1$, which implies than $AV \in \cV_1$. Thus $\cF_1^{(\theta)}$ satisfies inclusion condition, $A\circ \cF_1^{(\theta)} \subseteq \cF_1^{(\theta)}$.
%If $\maxnorm{A} = 1$, then  $\maxnorm{AV} \leq \maxnorm{A} \maxnorm{V} = 1 \cdot \maxnorm{V} = \max_{i} \sum_j |V_{ij}|$, which implies than $AV \in \cV_1$. Thus $\cF_1^{(\theta)}$ satisfies inclusion condition, $A\circ \cF_1^{(\theta)} \subseteq \cF_1^{(\theta)}$.
Since 
\begin{align}
\cV_2 = \big\{ \, v \in \reals^{d_v} \,\big|\, \|v\| \leq 1 \, \big\} = \big\{ \, \alpha v \,\big|\, v \in \reals^{d_v},\, \|v\| = 1 ,\, \alpha \in [0, 1] \, \big\}
\end{align}
we can re-write $\cF_2^{(\theta)}$ as,
\begin{align}
\cF_2^{(\theta)} = \bigg\{ \alpha f(x, y)  \,\,|\,\, f(x,y) = f(x), f(x) \in \cG^{(\theta, x)} = \{v^T\,\psi'(x;\theta) \,\big|\, \|v\| = 1 \}  , \alpha\in[0,1] \, \bigg\}\;.
\end{align}
Thus $\cF_2^{(\theta)}$ satisfies the label invariance condition. Finally, since Assumption \ref{eq:mix_inv} holds true for $\cF_1^{(\theta)}$ and $\cF_2^{(\theta)}$, it also holds for $\cF$.

% -------------------------------------------------------------------------------------------------------------------------------------------------------------------------
\section{Proof of Theorem \ref{thm:gen_counter}} \label{app:gen_counter_pf}
	Let $\cF = \{\D \,|\, \D(x) \in [-1, 1]^\ny \}$. We show that,
\begin{align*}
\cF_3 &= \left\lbrace \D \in \cF \, \bigg| \, \abs{\D(x)^T C^T (\vp(x) - \vq(x))} \leq \epsilon\; \forall x \in \cX \right\rbrace \,. \\
\cF_4 &= \left\lbrace \D \in \cF \, \bigg| \, \abs{\D(x)^T (\vp(x) - \vq(x))} \leq \epsilon\; \forall x \in \cX \right\rbrace
\end{align*}
Then,
\begin{align*}
\nndff{\cF_3}{\tP}{\tQ} &= \sup_{\D \in \cF} \;\; \expectD{(x,y) \sim \tP}{\D(x)_y} - \expectD{(x,y) \sim \tQ}{\D(x)_y} \\
&= \sup_{\D \in \cF} \;\; \expectD{(x,y) \sim P}{(C \D(x))_y} - \expectD{(x,y) \sim Q}{(C \D(x))_y} \\
&= \sup_{\D \in \cF} \;\; \int_\cX {\D(x)^T C^T  (p(x)\ygx{P}(y)) - q(x)\ygx{Q}(y))} \\
&\leq \sup_{\D \in \cF} \;\; \int_\cX \epsilon = O(\epsilon)
\\
\nndff{\cF_3}{P}{Q} &= \sup_{\D \in \cF} \;\; \expectD{(x,y) \sim P}{\D(x)_y} - \expectD{(x,y) \sim Q}{\D(x)_y} \\
&= \sup_{\D \in \cF} \;\; \int_\cX {\D(x)^T (p(x)\vygx{P} - q(x)\vygx{Q})} \\
&\geq \sup_{\D \in \cF} \;\; \int_{\cX_S} {\D(x)^T (p(x)\vygx{P} - q(x)\vygx{Q})} \\
&\overset{(a)}{\geq} \sup_{\substack{\D \in \cF \\ D(x) \perp v_x}} \;\; \int_{\cX_S}{\D(x)^T u_x}\\
&= \sup_{\D \in \cF} \;\; \int_{\cX_S}{\D(x)^T (u_x -  (u_x^T{\hat{v}_x})\hat{v}_x)} \\
&= \sup_{\D \in \cF} \;\; \int_{\cX_S}\onorm{u_x -  (u_x^T{\hat{v}_x})\hat{v}_x)},
\end{align*}
where in $(a)$ we put $p(x)\ygx{P}(y) - q(x)\ygx{Q}(y) = u_x$ and $\confus^T u_x = v_x$. Since $u_x$ is not and eigenvector of $\confus$, $u_x \not\perp v_x$ and therefore the integrand is positive.
Finally using the assumption that $P_X(\cX_S) + Q_X(\cX) > 0$ we get that LHS is positive number. Now by taking $\epsilon$ much smaller than the LHS we get the desired result. Other case also follows similarly.

% -------------------------------------------------------------------------------------------------------------------------------------------------------------------------
\section{Proof of Theorem \ref{thm:nn_sample}} \label{app:nn_sample_pf}
\begin{proposition} \label{pro:proj_disc}
	There exists a class $\cF$ of parametric vector valued functions which satisfy the Lipschitzness in parameters property \eqref{eq:param_lip}, such that $\cF - 1/2 \ones$ satisfies the inclusion condition (Assumption \ref{eq:mix_inv}).
	\begin{proof}
		For the proof we show that there exists a class of discriminators which satisfy the inclusion condition of Assumption \ref{eq:mix_inv} and in particular we have the following example. Let $\cF'$ be a class of vector functions parameterized by the $u' \in \cU'$ which is $L'$-Lipschitz in the parameters and who element functions satisfy $\maxnorm{f_{u'}(x)} \leq 1/2$. We define a new class of vector functions 
		\begin{align}
		\vcF \define \left\lbrace T' f_{u'}(\cdot) + 1/2\,\ones \,|\, \forall\; f(\cdot) \in \cF',\, \maxnorm{T'} \leq 1 \right\rbrace\,,
		\end{align}
		parameterized by $u = (u', T') \in \cU' \times \{T' \,|\, \maxnorm{T'} \leq 1\}$.
		Then,
		\begin{align}
		\maxnorm{T'_1 D_{u'_1}(x, \cdot) - T'_2 D_{u'_2}(x, \cdot)} 
		&\overset{(a)}{\leq} \maxnorm{T'_1 D_{u'_1}(x, \cdot) - T'_2 D_{u'_1}(x, \cdot)} +
		\maxnorm{T'_2 D_{u'_1}(x, \cdot) - T'_2 D_{u'_2}(x, \cdot)} \nonumber \\
		&\overset{(b)}{\leq} \tnorm{T'_1 D_{u'_1}(x, \cdot) - T'_2 D_{u'_1}(x, \cdot)} +
		\maxnorm{T'_2} \maxnorm{D_{u'_1}(x, \cdot) - D_{u'_2}(x, \cdot)} \nonumber \\
		&\overset{(c)}{\leq} \tnorm{T'_1 - T'_2}\tnorm{D_{u'_1}(x, \cdot)} + \maxnorm{T'_2} L' \tnorm{u'_1 - u'_2} \nonumber \\
		&\overset{(d)}{\leq} \tnorm{T'_1 - T'_2} \sqrt{\ny} + L' \tnorm{u'_1 - u'_2} \nonumber \\
		&\overset{}{\leq} (\sqrt{m} + L') \left(\tnorm{T'_1 - T'_2} + \tnorm{u'_1 - u'_2}\right) \nonumber\\
		&\overset{(e)}{\leq} \sqrt{2}(\sqrt{m} + L') \sqrt{\tnorm{T'_1 - T'_2}^2 + \tnorm{u'_1 - u'_2}^2}
		\end{align}
		where $(a)$ uses triangle inequality, $(b)$ and $(c)$ uses $\maxnorm{x} \leq \tnorm{x}$ and $\lnorm{Ax}{p} \leq \lnorm{A}{p} \lnorm{x}{p}$ (see Section \ref{sec:notations}), $(c)$ is true since $\vcF'$ is $L'$-Lipschitz in $u'$, $(d)$ uses $\maxnorm{D_{u'}(x, \cdot)} \leq 1/2$ and $\maxnorm{T_2'} \leq 1$, and $(e)$ uses $x + y \leq \sqrt{2 (x^2 + y^2)}$.
		
		Next we show that $\cF$ lies in the range $[0,1]^{\ny}$ as follows.
		\begin{align}
		\maxnorm{T' D_{u'}(x, \cdot)} \leq \maxnorm{T'} \maxnorm{D_{u'}(x, \cdot)} \overset{(b)}{\leq} 1 \cdot 1/2
		\end{align}
		We can prove inclusion condition in Assumption \ref{eq:mix_inv} by the fact that $\maxnorm{T T'} \leq \maxnorm{T} \maxnorm{T'} \leq 1 \cdot 1$ and hence $T T'$ is valid choice for for $T$.
	\end{proof}
\end{proposition}

Next we present a straightforward corollary of Theorem \ref{thm:gen_ub_lb}.
\begin{theorem} \label{thm:nn_ub_lb}
	Let $\cF$ be a parametric class of vector valued functions parameterized by $u \in \cU \subseteq \reals^p$ such that $f_u: \cX \to [0, 1]^m ,\; \forall f_u \in \cF$. Further, if $\cF$ is $L$-Lipschitz in the parameter $u$, as defined in equation \eqref{eq:param_lip}, and if $\cF$ satisfies Assumption $\ref{eq:mix_inv}$, then,
	\begin{align}
	\nndff{\cF}{\tP}{\tQ} \leq \nndff{\cF}{P}{Q} \leq \normconfusinv \nndff{\cF}{\tP}{\tQ}
	\end{align}
	\begin{proof}
		Proof directly follows from Theorem \ref{thm:gen_ub_lb}, since $T \circ \cF$ preserves Lipschitzness in $u$, when $\maxnorm{T} = 1$.
	\end{proof}
\end{theorem}

\begin{corollary}[of \citep{arora2017generalization}, Theorem 3.1.]
	\label{cor:nn_generalize}
	Assume the same class $\cF$ as in Theorem \ref{thm:nn_ub_lb}. Let $\tP$, $\tQ$ be two distributions on $X$, $\tY$ and be $\htP$, $\htQ$ be empirical versions of them with at least $n$ samples each. Then there is universal constant $c$ such that when $n \geq \frac{cp \log(p L/\epsilon)}{\epsilon^2}$, we have with probability at least $1 - exp(-p)$ over the randomness of $\htP$, $\htQ$,
	\begin{align}
	\abs{\nndff{\cF}{\htP}{\htQ} - \nndff{\cF}{\tP}{\tQ}} \leq \epsilon
	\end{align}
	\begin{proof}
		Proof is directly follow from \cite{arora2017generalization}[Theorem 3.1.].
	\end{proof}
\end{corollary}

\begin{alignat*}{2}
\nndff{\cF}{\tP}{\tQ} &\overset{(a)}\leq \nndff{\cF}{P}{Q} &&\overset{(a)}\leq \normconfusinv \nndff{\cF}{\tP}{\tQ} \\
\nndff{\cF}{\htP}{\htQ} - \epsilon &\overset{(a)}\leq \nndff{\cF}{P}{Q} &&\overset{(a)}\leq \normconfusinv (\nndff{\cF}{\htP}{\htQ} + \epsilon) \\
\end{alignat*}
%\begin{align}
%\nndff{\cF}{P}{Q} - \nndff{\cF}{\htP}{\htQ} &\overset{(a)}{\leq} \normconfusinv \nndff{\cF}{P}{Q} - \nndff{\cF}{\tP}{\tQ} \\
%&\overset{(b)}{\leq} \normconfusinv \epsilon + (\normconfusinv  -1)\nndff{\cF}{\tP}{\tQ} \\
%\end{align}
where $(a)$ is true by Theorem \ref{thm:nn_ub_lb} and $(b)$ is true from the Corollary \ref{cor:nn_generalize}.

% --------------------------------------------------------------------------------------------------------------------------------------------
\section{Implementation details}
\label{app:implementation}

{\bf Hyperparameters and architectures for MNIST:}  Biased GAN uses the standard conditional DCGAN architecture \cite{radford2015unsupervised} implementation\footnote{\url{https://github.com/carpedm20/DCGAN-tensorflow}}. RCGAN, RCGAN-U, and unbiased GAN use the same DCGAN architecture \cite{radford2015unsupervised} with hinge loss, $\phi(a)= \max(0,1-2a)$, and conditional projection discriminator \cite{miyato2018cgans}, as suggested by our theoretical analysis. Additionally, we also present RCGAN+y architecture, which has the same architecture as RCGAN but the input to the first layer of its discriminator is concatenated with a one-hot representation of the label. 

We use $\lambda=0$ and $\lambda=1$ for RCGAN and RCGAN-U respectively, and a linear classifier for the permutation regularizer. For RCGAN-U, the learning rate of the confusion matrix is $10$ times as that of the discriminator and the generator.

\bigskip\noindent
{\bf RCGAN+y training:} RCGAN+y architecture has the same architecture as RCGAN but the input to the first layer of its discriminator is concatenated with a one-hot representation of the label. We observed that RCGAN+y is harder to train especially at low noise regimes of $\alpha \geq 0.4$. To combat this we add additional artificial noise, parameterized by $\tilde{\alpha}$, so that the effective noise parameter is $\bar{\alpha} = \tilde{\alpha} (\alpha - \frac{(1-\alpha)}9) + \frac{(1-\alpha)}9$. We schedule $\tilde\alpha$ during the training so that from epoch $0$ to $30$, the effective noise $\bar{\alpha} = 0.3$. When the original noise parameter $\alpha$ value is $0.9$, from epoch $30$ to $80$, we linearly increase $\tilde{\alpha}$ so that the effective noise parameter linearly decrease from $0.3$ to $\alpha$, and from epoch $80$ to $100$ we keep the artificial noise parameter $\tilde\alpha$ to $1$, so that effective noise parameter $\bar{\alpha} = \alpha$. For original noise parameter $\alpha$ value less than $0.9$, starting from epoch $30$ we increase $\tilde\alpha$ in the same rate as in the case of $\alpha=0.9$ and once it reaches $1$ keep it constant till the end of training.
This  stabilizes the training of RCGAN+y. 
%But the same process could not stabilize the training for this architecture when confusion matrix was unknown and had to learned. 

We suspect that the reason why RCGAN+y works better than RCGAN is that 
RCGAN+y optimizes over a larger class of functions, 
which might be necessary when learning a conditional distribution with large noise.  
%as noise increases the real noisy distribution $\tP$ and fake noisy distribution $\tQ$ 
%becomes highly complicated in $y$. 
Thus, as the noise increases it becomes more challenging for the projection discriminator to differentiate between $\tP$ and $\tQ$, 
since it is much simpler function of $y$ with $y$ appearing only in the final layer. 
However, 
 in RCGAN+y since we concatenate $y$ to the input of the first layer, RCGAN+y discriminator may be able to better differentiate between $\tP$ and $\tQ$. 

\bigskip\noindent
{\bf Hyperparameters and architectures for CIFAR-$10$:} We use ResNet based GAN used in \cite{gulrajani2017improved} with spectral normalization \cite{miyato2018spectral} and conditional projection discriminator \cite{miyato2018cgans} \footnote{\url{https://github.com/watsonyanghx/GAN\_Lib\_Tensorflow}}.  We note that the spectral normalization work \cite{miyato2018spectral} reports higher inception score than what we achieved on the noiseless setting, possibly due to limited hyper-parameter tuning. For all the four approaches we use the same hyper-parameters. 

We use $\lambda=0$ and $\lambda=1$ for RCGAN and RCGAN-U respectively, and a linear classifier for the permutation regularizer. For RCGAN-U, the learning rate of the confusion matrix is same as that of the discriminator and the generator.

\bigskip\noindent
{\bf RCGAN-U CIFAR-$10$ initialization:} For CIFAR-$10$ dataset, we observed that even with the permutation regularizer, the learned confusion matrix in RCGAN-U was a permuted version of the true $C$, possibly because a linear classifier might be too simple to classify CIFAR images. To combat this, we initialized the confusion matrix $M$ to be diagonally dominant. We initialized the confusion matrix to be learned $M$ so that diagonal entries are $0.2$ and off-diagonal entries are $(1-0.2)/9$. In our experiments, this ensured that the approximately true confusion matrix $C$ was learned by $M$ in RCGAN-U. We believe that better CIFAR-$10$ classifier for permutation regularizers can achieve the same effect as this initialization.

% ---------------

\section{Additional experimental information}
\label{app:A}

In this section, we provide tables with the numerical values of the data points in the Figures \ref{fig:mnist1} and \ref{fig:cifar1}.

\begin{table}[h]
\begin{center}
\begin{tabular}{ c c c c c c}
  {\bf Noise ($1-\alpha$)} & {\bf RCGAN+y} & {\bf RCGAN} & {\bf RCGAN-U} & {\bf unbiased GAN} & {\bf biased GAN} \\\hline
   0.875  &	0.905 	& 0.11 	& 0.215 & 0.1 	& 0.119 \\
   0.85   &	0.984 	& 0.211 & 0.235 & 0.1 	& 0.138 \\
   0.8    &	0.987 	& 0.44 	& 0.489 & 0.1 	& 0.192 \\
   0.7    &	0.985 	& 0.978 & 0.992 & 0.1 	& 0.288 \\   
   0.6    &	0.976 	& 0.983 & 0.991 & 0.2 	& 0.375 \\   
   0.5    &	0.976 	& 0.991 & 0.986 & 0.869 & 0.475 \\   
   0.4    &	0.984 	& 0.994 & 0.99 	& 0.997 & 0.57 \\   
   0.3    &	0.983 	& 0.994 & 0.987 & 0.991 & 0.681 \\   
   0.2    &	0.99 	& 0.994 & 0.995 & 0.999 & 0.765 \\   
   0.1    &	0.99 	& 0.994 & 0.995 & 0.998 & 0.873 \\   
   0.0    &	0.995 	& 0.995 & 0.994 & 0.994 & 0.994 \\   

\end{tabular}
\caption{Numerical values for the datapoints in Figure \ref{fig:mnist1} (left panel). Noisy MNIST dataset: Generated label accuracy of RCGAN, RCGAN-U, RCGAN+y, unbiased GAN and biased GAN.}
\label{tab:mnist_gen_label_acc}
\end{center}
\end{table}

\begin{table}[h]
\begin{center}
\begin{tabular}{ c c c c c c}
  {\bf Noise ($1-\alpha$)} & {\bf RCGAN+y} & {\bf RCGAN} & {\bf RCGAN-U} & {\bf unbiased GAN} & {\bf biased GAN} \\\hline
   0.875  &	0.156 	& 0.885 & 0.86	& 0.898 & 0.872 \\
   0.85   &	0.102 	& 0.774 & 0.77	& 0.894	& 0.854 \\
   0.8    &	0.088 	& 0.638	& 0.69	& 0.9 	& 0.634 \\
   0.7    &	0.11 	& 0.096 & 0.098 & 0.768	& 0.55 \\   
   0.6    &	0.088 	& 0.1	& 0.058 & 0.902	& 0.322 \\   
   0.5    &	0.07 	& 0.106 & 0.094 & 0.472	& 0.274 \\   
   0.4    &	0.072 	& 0.098 & 0.08 	& 0.158	& 0.164 \\   
   0.3    &	0.096 	& 0.088 & 0.084 & 0.098	& 0.142 \\   
   0.2    &	0.076 	& 0.086 & 0.086 & 0.07 	& 0.138 \\   
   0.1    &	0.112 	& 0.068 & 0.096 & 0.088	& 0.104 \\   
   0.0    &	0.069 	& 0.069 & 0.069 & 0.069	& 0.069 \\   

\end{tabular}
\caption{Numerical values for the datapoints in Figure \ref{fig:mnist1} (right panel).
 Noisy MNIST dataset: Label recovery error of RCGAN, RCGAN-U, RCGAN+y, unbiased GAN and biased GAN.}
\label{tab:mnist_label_recovery_acc}
\end{center}
\end{table}

\begin{table}[h]
\begin{center}
\begin{tabular}{ c c c c c }
  {\bf Noise ($1-\alpha$)} & {\bf RCGAN} & {\bf RCGAN-U} & {\bf unbiased GAN} & {\bf biased GAN} \\\hline
   0.8   & 0.111	& 0.126 & 0.110 & 0.117 \\
   0.6   & 0.443	& 0.263 & 0.1 	& 0.148 \\   
   0.4   & 0.700	& 0.71 	& 0.4 	& 0.340 \\   
   0.2   & 0.724	& 0.815 & 0.6 	& 0.507 \\   
   0.0   & 0.751	& 0.751 & 0.751 & 0.751 \\   
\end{tabular}
\caption{Numerical values for the datapoints in Figure \ref{fig:cifar1} (left panel). Noisy CIFAR-$10$ dataset: Generated label accuracy of RCGAN, RCGAN-U, unbiased GAN and biased GAN.}
\label{tab:cifar_gen_label_acc}
\end{center}
\end{table}

\begin{table}[h]
\begin{center}
\begin{tabular}{ c c c c c }
  {\bf Noise ($1-\alpha$)} & {\bf RCGAN} & {\bf RCGAN-U} & {\bf unbiased GAN} & {\bf biased GAN} \\\hline
   0.8   & 7.8		& 7.58 	& 4.37 	& 7.6 \\
   0.6   & 7.85		& 7.56	& 4 	& 7.68 \\   
   0.4   & 8.05		& 8.03 	& 4		& 7.75 \\   
   0.2   & 8.11		& 8.12	& 6		& 7.91 \\   
   0.0   & 8.13		& 8.13 	& 8.13 	& 8.13 \\ 
\end{tabular}
\caption{Numerical values for the datapoints in Figure \ref{fig:cifar1} (right panel). Noisy CIFAR-$10$ dataset: Inception score of RCGAN, RCGAN-U, unbiased GAN and biased GAN.}
\label{tab:CIFAR_label_recovery_acc}
\end{center}
\end{table}

\def\sampleparent{}
\begin{figure}[ht]
	\def\alpha{0.2}
	\def\algo{rcgany}
	\foreach \alpha in {0.2,0.3,0.4,0.5}
	{
		\includegraphics[width=0.18\textwidth]{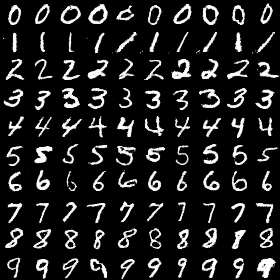}	
		\hspace{0.25em} 
	}
	\put(-370,95){label accuracy 0.2}
	\put(-273,95){label accuracy 0.3}
	\put(-180,95){label accuracy 0.4}
	\put(-80,95){label accuracy 0.5}
	\put(-395,20){\rotatebox{90}{\small{RCGAN+y}}}
	
	\vspace{0.25em}
	
	\def\alpha{0.2}
	\def\algo{rcgan}
	\foreach \alpha in {0.2,0.3,0.4,0.5}
	{
		\includegraphics[width=0.18\textwidth]{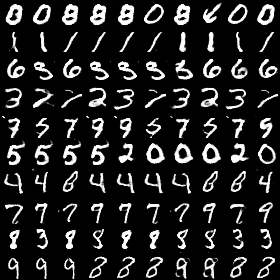}	
		\hspace{0.25em} 
	}
	\put(-395,20){\rotatebox{90}{\small{RCGAN}}}
	\vspace{0.25em}	
	
	\def\alpha{0.2}
	\def\algo{rcganu}
	\foreach \alpha in {0.2,0.3,0.4,0.5}
	{
		\includegraphics[width=0.18\textwidth]{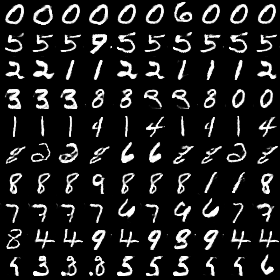}	
		\hspace{0.25em} 
	}
	\put(-395,20){\rotatebox{90}{\small{RCGAN-U}}}
	\vspace{0.25em}
	
	\def\alpha{0.2}
	\def\algo{unbias}
	\foreach \alpha in {0.2,0.3,0.4,0.5}
	{
		\includegraphics[width=0.18\textwidth]{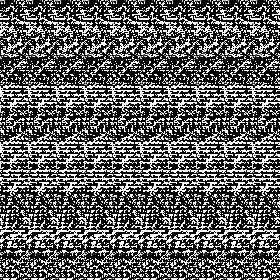}	
		\hspace{0.25em} 
	}
	\put(-395,20){\rotatebox{90}{\small{Unbiased}}}
	\vspace{0.25em}	
	
	\def\alpha{0.2}
	\def\algo{bias}
	\foreach \alpha in {0.2,0.3,0.4,0.5}
	{
		\includegraphics[width=0.18\textwidth]{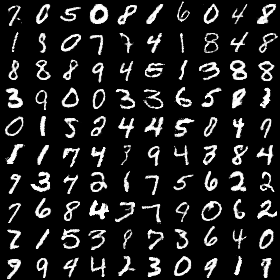}	
		\hspace{0.25em} 
	}
	\put(-395,20){\rotatebox{90}{\small{Biased}}}
	\vspace{0.75em}
		
	\def\alpha{1.0}
	\centering\includegraphics[width=0.18\textwidth]{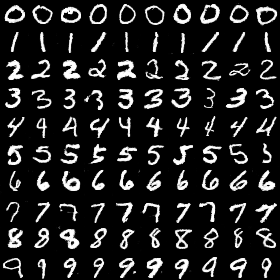}
	\put(-100,6){\rotatebox{90}{\small{label accuracy $\alpha$}}}
	\caption{Images generated from the RCGAN+y, RCGAN , RCGAN-U, Unbiased and Biased GANs trained using noisy MNIST dataset, where class labels are flipped uniformly at random with probability $1-\text{\rm{`real label accuracy'}}$, under the uniform flipping model.}
	\label{fig:mnist3}
\end{figure}

\begin{figure}[ht]
	\def\alpha{0.2}
	\includegraphics[width=0.18\textwidth]{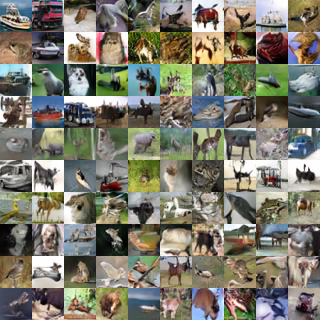}
	\hspace{0.5em}
	\includegraphics[width=0.18\textwidth]{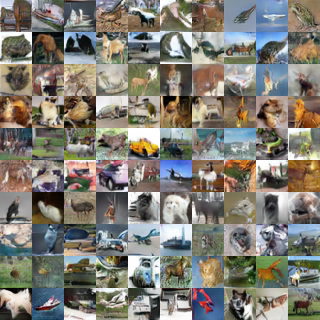}
	\hspace{0.5em} 
	\includegraphics[width=0.18\textwidth]{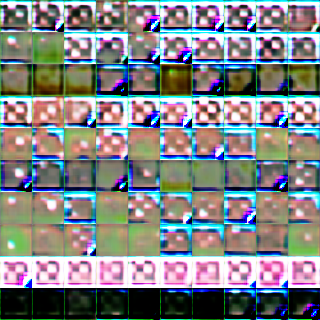}
	\hspace{0.5em} 	
	\includegraphics[width=0.18\textwidth]{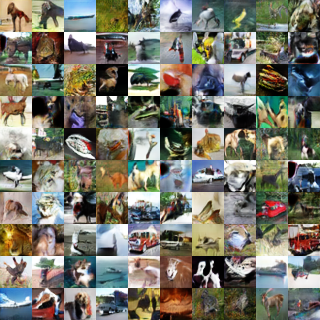}
	\put(-360,95){RCGAN-U}
	\put(-260,95){RCGAN}
	\put(-170,95){Unbiased}
	\put(-60,95){Biased}
	\put(-400,12){\rotatebox{90}{\small{label accuracy $\alpha$}}}
	\\
	\foreach \alpha in {0.4,0.6,0.8}
	{
		\includegraphics[width=0.18\textwidth]{\sampleparent./rcganu_\alpha_samples_24999.png}
		\hspace{0.5em}
		\includegraphics[width=0.18\textwidth]{\sampleparent./rcgan_\alpha_samples_31999.png}
		\hspace{0.5em} 
		\includegraphics[width=0.18\textwidth]{\sampleparent./unbiased_\alpha_samples_31999.png}
		\hspace{0.5em} 	
		\includegraphics[width=0.18\textwidth]{\sampleparent./biased_\alpha_samples_31999.png}
		\put(-400,12){\rotatebox{90}{\small{label accuracy $\alpha$}}}\\
	}	
	\vspace{0.75em}	
	\def\alpha{1.0}
	\centering\includegraphics[width=0.18\textwidth]{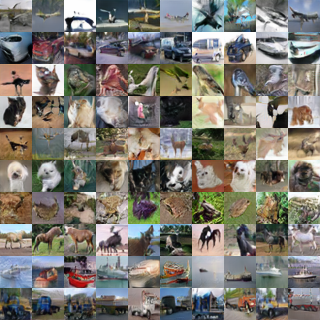}
	\put(-110,10){\rotatebox{90}{\small{label accuracy $\alpha$}}}
	\caption{Images generated from the RCGAN-U, Unbiased and Biased GANs trained using noisy CIFAR-$10$ dataset, where class labels are flipped uniformly at random with probability $1-\text{\rm{`real label accuracy'}}$, under the uniform flipping model.}
	\label{fig:cifar2}
\end{figure}

\end{document}